\documentclass{article}
\usepackage{log_2022}         % for camera-ready version
\usepackage{booktabs}           % professional-quality tables
\usepackage{multirow}           % tabular cells spanning multiple rows
\usepackage{amsfonts}           % blackboard math symbols
\usepackage{graphicx}           % figures
\usepackage{duckuments}         % sample images
\usepackage[numbers,compress,sort]{natbib}
\usepackage{caption}
\usepackage{subcaption}
\usepackage{algorithm}
\usepackage{algorithmic}
\renewcommand{\algorithmicrequire}{\textbf{Input:}}
\renewcommand{\algorithmicensure}{\textbf{Output:}}
\usepackage{amssymb}

\usepackage{amsthm}
\newtheorem{definition}{Definition}
\title{\textcolor{black}{Transductive} Linear Probing: A Novel \textcolor{black}{Framework} \\for Few-Shot Node Classification}

\author[Z. Tan et al.]{%
Zhen Tan\thanks{Equal contribution.}\\
\institute{Arizona State University}\\
\email{ztan36@asu.edu}\And
\hspace{0.4in}Song Wang\footnotemark[1]\\
\hspace{0.4in}\institute{University of Virginia}\\
\hspace{0.4in}\email{sw3wv@virginia.edu}\And
\hspace{0.4in}Kaize Ding\footnotemark[1]\\
\hspace{0.4in}\institute{Arizona State University}\\
\hspace{0.4in}\email{kding9@asu.edu}
\And
Jundong Li\\
\institute{University of Virginia}\\
\email{jundong@virginia.edu}\And
Huan Liu\\
\institute{Arizona State University}\\
\email{huanliu@asu.edu}
}

\begin{document}
\maketitle
\begin{abstract}
Few-shot node classification is tasked to provide accurate predictions for nodes from novel classes with only few representative labeled nodes. This problem has drawn tremendous attention for its projection to prevailing real-world applications, such as product categorization for newly added commodity categories on an E-commerce platform with scarce records or diagnoses for rare diseases on a patient similarity graph. To tackle such challenging label scarcity issues in the non-Euclidean graph domain, meta-learning has become a successful and predominant paradigm.
More recently, inspired by the development of graph self-supervised learning, transferring pretrained node embeddings for few-shot node classification could be a promising alternative to meta-learning but remains unexposed. In this work, we empirically demonstrate the potential of an alternative framework, \textit{Transductive Linear Probing}, that transfers pretrained node embeddings, which are learned from graph contrastive learning methods. We further extend the setting of few-shot node classification from standard fully supervised to a more realistic self-supervised setting, where meta-learning methods cannot be easily deployed due to the shortage of supervision from training classes. Surprisingly, even without any ground-truth labels, transductive linear probing with self-supervised graph contrastive pretraining can outperform the state-of-the-art fully supervised meta-learning based methods under the same protocol. We hope this work can shed new light on few-shot node classification problems and foster future research on learning from scarcely labeled instances on graphs.
\end{abstract}

\section{Introduction}
\label{sec:intro}
Graph Neural Networks (GNNs) \cite{Kipf:2017tc,Velickovic:2018we,Hamilton:2017tp,xu2018powerful} are a family of neural network models designed for graph-structured data. In this work, we concentrate on GNNs for the node classification task, where GNNs recurrently aggregate neighborhoods to simultaneously preserve graph structure information and learn node representations. However, most GNN models focus on the (semi-)supervised learning setting, assuming access to abundant labels~\cite{ding2022meta,ding2022data}. This assumption could be practically infeasible due to the high cost of data collection and labeling, especially for large graphs. Moreover, recent works have manifested that directly training GNNs with limited nodes can result in severe performance degradation \cite{zhang2018few,ding2020graph,wang21AMM}. Such a challenge has led to a proliferation of studies \cite{huang2020graph,lan2020node,wang2022task,ding2022data} that try to learn fast-adaptable GNNs with extremely scarce known labels, \textit{i.e., Few-Shot Node Classification} (FSNC) tasks. Particularly, in FSNC, there exist two disjoint label spaces: \textit{base classes} are assumed to contain substantial labeled nodes while target \textit{novel classes} only contain few available labeled nodes. If the target FSNC task contains $N$ novel classes with $K$ labeled nodes in each class, the problem is 
denoted as an $N$-way $K$-shot node classification task. Here the $K$ labeled nodes are termed as a \textit{support set}, and the unlabeled nodes are termed as a \textit{query set} for evaluation.

Currently, \textit{meta-learning} has become a prevailing and successful paradigm to tackle such a shortage of labels on graphs. Inspired by the way humans learn unseen classes with few samples
via utilizing previously learned prior knowledge, a typical meta-learning based framework will randomly sample a number of \textit{episodes}, or \textit{meta-tasks}, to emulate the target $N$-way $K$-shot setting~\cite{zhang2018few}. Based on this principle, various models~\cite{zhang2018few,ding2020graph,wang21AMM,huang2020graph,lan2020node,wang2022task,zhang2022few} have been proposed, which makes meta-learning a plausible default choice for FSNC tasks. 
% However, to the best of our knowledge, there is no unified evaluation protocol. The existence of these different models each evaluated with their own protocols and datasets fragments the practical knowledge on how meta-learning performs with a few labeled nodes and hinders the explicit comparison between models. This gap engenders our primary motivation to offer a low barrier of entry to compare new advances and prior works for FSNC tasks. 
On the other hand, despite the remarkable breakthroughs that have been made, meta-learning based methods still have several limitations. \textbf{First}, relying on different arbitrarily sampled meta-tasks to extract transferable meta-knowledge, meta-learning based frameworks suffer from the piecemeal knowledge issue~\cite{tansupervised}. That being said, a small portion of the nodes and classes are selected per episode for training, which leads to an undesired loss of generalizability of the learned GNNs regarding nodes from unseen novel classes. \textbf{Second}, the feasibility for sampling meta-tasks is based on the assumption that there exist sufficient base classes where substantial labeled nodes are accessible. However, this assumption can be easily overturned for real-world graphs where the number of base classes can be limited, or the labels of nodes in base classes can be inaccessible. In a nutshell, these two concerns motivate us to design an alternative \textcolor{black}{framework} for meta-learning to cover more realistic scenarios.

% Thus, in this paper, we extend the FSNC problem setting to cover those scenarios and accordingly provide an alternative to meta-learning. 

Inspired by \cite{dhillon2019baseline,tian2020rethinking}, we postulate that the key to solving FSNC
is to learn a generalizable GNN encoder. We validate this postulation by a motivating example in Section \ref{sec:motivate}. Then, without the episodic emulation, the proposed novel framework, \textit{Transductive Linear Probing} (TLP), directly transfers pretrained node embeddings for nodes in \textit{novel classes} learned from \textit{Graph Contrastive Learning} (GCL) methods~\cite{hassani2020contrastive,you2020graph,zhu2020deep,jin2021multi,thakoor2021large,mo2022simple,ding2023structural}, and fine-tunes a separate linear classifier with the support set to predict labels for unlabeled nodes. GCL methods are proven to learn generalizable node embeddings by maximizing the representation consistency under different augmented views \cite{chen2020simple,hassani2020contrastive,you2020graph,ding2023structural}. If the representations of nodes in novel classes are discriminative enough, probing them with a simple linear classifier should provide decent accuracy. Based on this intuition, we propose two instantiations of the TLP framework in this paper: TLP with the self-supervised form of GCL methods and TLP with the supervised GCL counterparts. We evaluate TLP by transferring node embeddings from various GCL methods to the linear classifier and compare TLP with meta-learning based methods under the same evaluation protocol. Moreover, we examine the effect of supervision during GCL pretraining for target FSNC tasks to further analyze what role labels from base classes play in TLP.

Throughout this paper, we aim to shed new light on the few-shot node classification problem through the lens of empirical evaluations of both the "old" meta-learning paradigm and the "new" transductive linear probing framework. The summary of our contributions is as follows: 

\textbf{New Framework} We are the first to break with convention and precedent to propose a new framework, transductive linear probing, as a competitive alternative to meta-learning for FSNC tasks.

\textbf{Comprehensive Study} We perform comprehensive reviews on current literature and the research community and conduct a large-scale study on six widely-used real-world datasets that cover different scenarios in FSNC: (1) a sufficient number of base classes with substantial labeled nodes in each class, (2) a sufficient number of base classes with no labeled nodes in each class, (3) a limited number of base classes with substantial labeled nodes in each class, and (4) a limited number of base classes with no labeled nodes in each class. We evaluate all the compared methods under the same protocol.

\textbf{Findings} We demonstrate that despite the recent advances in few-shot node classification, meta-learning based methods struggle to outperform TLP methods. Moreover, the TLP-based methods with self-supervised GCL can outperform their supervised counterparts and those meta-learning based methods even if all the labels from base classes are inaccessible. This signifies that without label information, self-supervised GCL can focus more on node-level structural information, which results in better node representations. However, TLP also inherits its limitation for scalability due to the large memory consumption of GCL, which makes it hard to deploy on extremely large graphs. Based on those observations, we identify that improving adaptability and scalability are the promising directions for meta-learning based and TLP-based methods, respectively.

Our implementations for experiments are released\footnote{\url{https://github.com/Zhen-Tan-dmml/TLP-FSNC.git}}. We hope to facilitate the sharing of insights and accelerate the progress on the goal of learning from scarcely labeled instances on graphs.

\section{Preliminaries}
\label{sec:Prel}
\subsection{Problem Statement}
\label{sec:problem}
Formally, given an attributed network $\mathcal{G} = (\mathcal{V}, \mathcal{E}, \mathbf{X}) = (\mathbf{A}, \mathbf{X})$, where $\mathcal{V}$ denotes the set of nodes $\{v_1, v_2, ..., v_n\}$, $\mathcal{E}$ denotes the set of edges $\{e_1, e_2, ..., e_m\}$, $\mathbf{X} = [\mathbf{x}_1;\mathbf{x}_2; ...;\mathbf{x}_n] \in \mathbb{R}^{n\times d}$ denotes all the node features, and $\mathbf{A} = \{0, 1\}^{n\times n}$ is the adjacency matrix representing the network structure. Specifically, $\mathbf{A}_{j,k} = 1$ indicates that there is an edge between node $v_j$ and
node $v_k$; otherwise, $\mathbf{A}_{j,k} = 0$. The few-shot node classification problem assumes that there exist a series of target node classification tasks, $\mathcal{T} = \{\mathcal{T}_i\}^{I}_{i=1}$, where $\mathcal{T}_i$ denotes the given dataset of a task, and $I$ denotes the number of such tasks.
% $\mathbf{X}_{\mathbb{C}^{i}}$ denotes the attributes of nodes whose labels belong to the label space $\mathbb{C}^{i}$, and $\mathbf{A}_{\mathbb{C}^{i}}$ similarly. 
We term the classes of nodes available during training as base classes (i.e., $\mathbb{C}_{base}$) and the classes of nodes during target test phase as novel classes (i.e., $\mathbb{C}_{novel}$) and $\mathbb{C}_{base} \cap \mathbb{C}_{novel} = \varnothing$. Notably, under different settings, labels of nodes for training (i.e., $\mathbb{C}_{base}$) may or may not be available during training.
% , and the number of base classes, $|\mathbb{C}_{base}|$, is not necessarily large. 
Conventionally, there are few labeled nodes for novel classes $\mathbb{C}_{novel}$ during the test phase. The problem of few-shot node classification is defined as follows:

\begin{definition}
\textbf{Few-shot Node Classification:} Given an attributed graph $\mathcal{G} = (\mathbf{A}, \mathbf{X})$ with a divided node label space $\mathbb{C} = \{\mathbb{C}_{base}, \mathbb{C}_{novel}\}$, we only have few-shot labeled nodes (support set $\mathbb{S}$) for $\mathbb{C}_{novel}$. The task $\mathcal{T}$ is to predict the labels for unlabeled nodes (query set $\mathbb{Q}$) from $\mathbb{C}_{novel}$. If the support set in each target (test) task has $N$ novel classes with $K$ labeled nodes, then we term this task an $N$-way $K$-shot node classification task.
\end{definition}

The goal of few-shot node classification is to learn an encoder that can transfer the topological and semantic knowledge learned from substantial data in base classes ($\mathbb{C}_{base}$) and generate discriminative embeddings for nodes from novel classes ($\mathbb{C}_{novel}$) with limited labeled nodes.

\subsection{Episodic Meta-learning for Few-shot Node Classification.}

Episodic meta-learning is a proven effective paradigm for few-shot learning tasks \cite{mishra2018simple,ravi2016optimization,nichol2018first,liu2019learning,sung2018learning,snell2017prototypical,finn2017model,wang2022faith}. The main idea is to train the neural networks in a way that emulates the evaluation conditions. This is hypothesized to be beneficial for the prediction performance on test tasks \cite{mishra2018simple,ravi2016optimization,wang2021reform,nichol2018first}. Based on this philosophy, many recent works in few-shot node classification  \cite{zhou2019meta,ding2020graph,lan2020node,huang2020graph,liu2021relative,tan2022graph,liu2022few,wang2022task,wuinformation,wang2022extreme} successfully transfer the idea to the graph domain. 
It works as follows: during the training phase, it generates a number of meta-train tasks (or episodes) $\mathcal{T}_{tr}$ from $\mathbb{C}_{base}$ to emulate the test tasks, following their $N$-way $K$-shot node classification specifications:
\begin{align}
\begin{aligned}
    \mathcal{T}_{tr} &= \{\mathcal{T}_t\}_{t=1}^T = \{\mathcal{T}_1, \mathcal{T}_2, ..., \mathcal{T}_T\}, \\
    \mathcal{T}_t &= \{\mathcal{S}_t, \mathcal{Q}_t\}, \\
    \mathcal{S}_t &= \{(v_1, y_1), (v_2, y_2), ..., (v_{N\times K}, y_{N\times K})\}, \\
    \mathcal{Q}_t &= \{(v_1, y_1), (v_2, y_2), ..., (v_{N\times K}, y_{N\times K})\}. 
\end{aligned}
\end{align}
For a typical meta-learning based method, in each episode, $K$ labeled nodes are randomly sampled from $N$ base classes, forming a \textit{support set}, to train the GNN model while emulating the $N$-way $K$-shot node classification in the test phase. Then GNN predicts labels for an emulated \textit{query set} of nodes randomly sampled from the same classes as the support set. The Cross-Entropy Loss ($L_{CE}$) is calculated to optimize the GNN encoder $g_\theta$ and the classifier $f_\psi$ in an end-to-end fashion:
\begin{equation}
    \theta, \psi = \arg\min_{\theta, \psi} L_{CE}(\mathcal{T}_{t};\theta, \psi).
\end{equation}
Based on this, Meta-GNN~\cite{zhou2019meta} combines MAML~\cite{finn2017model} with GNNs to achieve optimization for different meta-tasks. GPN~\cite{ding2020graph} applies ProtoNet~\cite{snell2017prototypical} and computes node importance for a transferable metric function. G-Meta~\cite{huang2020graph} aims to establish a local subgraph for each node to achieve fast adaptations to new meta-tasks. RALE~\cite{liu2021relative} obtains relative and absolute node embeddings based on node positions on graphs to model node dependencies in each meta-task. An exhaustive survey is beyond the scope of
this paper; see \cite{zhang2022few} for an overview. However, all those methods are evaluated on different datasets with each own evaluation protocol, which fragments the practical knowledge on how meta-learning performs with a few labeled nodes and makes it hard to explicitly compare their superiority or inferiority. To bridge this gap, in this paper, we conduct extensive experiments to compare new advances and prior works for FSNC tasks uniformly and comprehensively.

% Few-shot learning targets at achieving considerable performance using only a limited number of labeled samples as references. In general, the prevalent approach is to learn transferable knowledge from existing tasks with abundant labeled samples and then generalize it to novel tasks with few labeled samples. 

% Typically, existing few-shot learning methods can be divided as two main categories: (1) \emph{Metric-based} approaches propose to learn a metric function to measure the similarity between the query set and the support set for classification~\cite{liu2019learning,sung2018learning}. As an example, Matching Networks~\cite{vinyals2016matching} leverage a nearest-neighbor matching strategy for few-shot classification. Prototypical Networks~\cite{snell2017prototypical} compute a prototype for each class and conduct classification according to the Euclidean distances between the query set and the prototypes. (2) \emph{Optimization-based} approaches aim to learn optimal model parameters via gradients on support samples in each meta-task~\cite{mishra2018simple,ravi2016optimization,nichol2018first}. As a classic example, MAML~\cite{finn2017model} learns an optimal parameter initialization for different meta-tasks with the proposed meta-optimization strategy. LSTM-based meta-learner~\cite{ravi2016optimization} proposes an adjustable step size to update model parameters. In addition, to obtain an optimal learning strategy.

\subsection{A Motivating Example and Preliminary Analysis}\label{sec:motivate}

More recently, related works in the image domain demonstrate that the reason for the fast adaptation lies in feature reuse rather than those complicated mate-learning algorithms~\cite{dhillon2019baseline,tian2020rethinking}. In other words, with a carefully pretrained encoder, decent performance can be obtained through directly fine-tuning a simple classifier on the target task.
% Since then, various pretraining strategies, especially contrastive based pretraining strategies \cite{liu2021learning,gao2021contrastive,ouali2021spatial}, have been put forward to tackle the few-shot image classification problem. 
    However, few studies have been done on the graph domain due to its important difference from images that nodes in a graph are not i.i.d. Their interactive relationships are reflected by both the topological and semantic information. To validate such hypothesis on graphs, based on \cite{tian2020rethinking}, we construct an \textit{Intransigent GNN} model, namely \textit{I-GNN}, that simply does not adapt to new tasks. We decouple the training procedure to two separate phases. In the first phase, a GNN encoder $g_\theta$ with a linear classifier $f_{\phi}$ as the classifier is simply pretrained on all base classes $\mathbb{C}_{base}$ with vanilla supervision through $L_{CE}$:
\begin{equation}
\label{eq:intransigent}
\begin{aligned}
    \mathcal{T}_{tr}^\prime &= \cup\{\mathcal{T}_t\}_{t=1}^T = \cup\{\mathcal{T}_1, \mathcal{T}_2, ..., \mathcal{T}_T\}, \\
    \theta, \phi &= \arg\min_{\theta, \phi} L_{CE}(\mathcal{T}_{tr}^\prime;\theta, \phi) + \mathcal{R}(\theta),
\end{aligned}
\end{equation}

where $\mathcal{R}(\theta)$ is a weight-decay regularization term: $\mathcal{R}(\theta) = \|\theta\|^2 /2$. Then, we freeze the parameter of the GNN encoder $g_\theta$ and discard the classifier $f_\phi$. When fine-tuning on a target few-shot node classification task $\mathcal{T}_i = \{\mathcal{S}_i, \mathcal{Q}_i\}$, the embeddings of all nodes from $\mathcal{T}_i$ are directly transferred from the pretrained GNN encoder $g_\theta$. Then another linear classifier $f_\psi$ is involved and tuned with few-shot labeled nodes from the support set $\mathcal{S}_i$ to predict labels of nodes in the query set $\mathcal{Q}_i$:
\begin{equation}
    \label{eq:ft}
    \psi = \arg\min_\psi L_{CE}(\mathcal{S}_i;\theta,\psi).
\end{equation}
\paragraph{Results and Analysis of the Intransigent GNN model I-GNN}
We demonstrate the performance of the intransigent model and compare it with those meta-learning based models in Table~\ref{tab:all_result}, \ref{tab:otherdata}. Under the same evaluation protocol (defined in Section \ref{sec:protocol}), the simple intransigent model I-GNN has very competitive performance with meta-learning based methods. On datasets (e.g., \texttt{CiteSeer}) where the number of base classes $|\mathbb{C}_{base}|$ is limited, I-GNN consistently outperforms meta-learning based methods in terms of accuracy. This motivating example concludes that transferring node embeddings from the vanilla supervised training method I-GNN could be an alternative to meta-learning. Moreover, we take one step further and postulate that if more transferable node embeddings are obtained during pretraining, the performance on target FSNC tasks could be improved even more. 

\begin{figure}[htbp]
  \centering\scalebox{0.99}{
  \includegraphics[width=\linewidth]{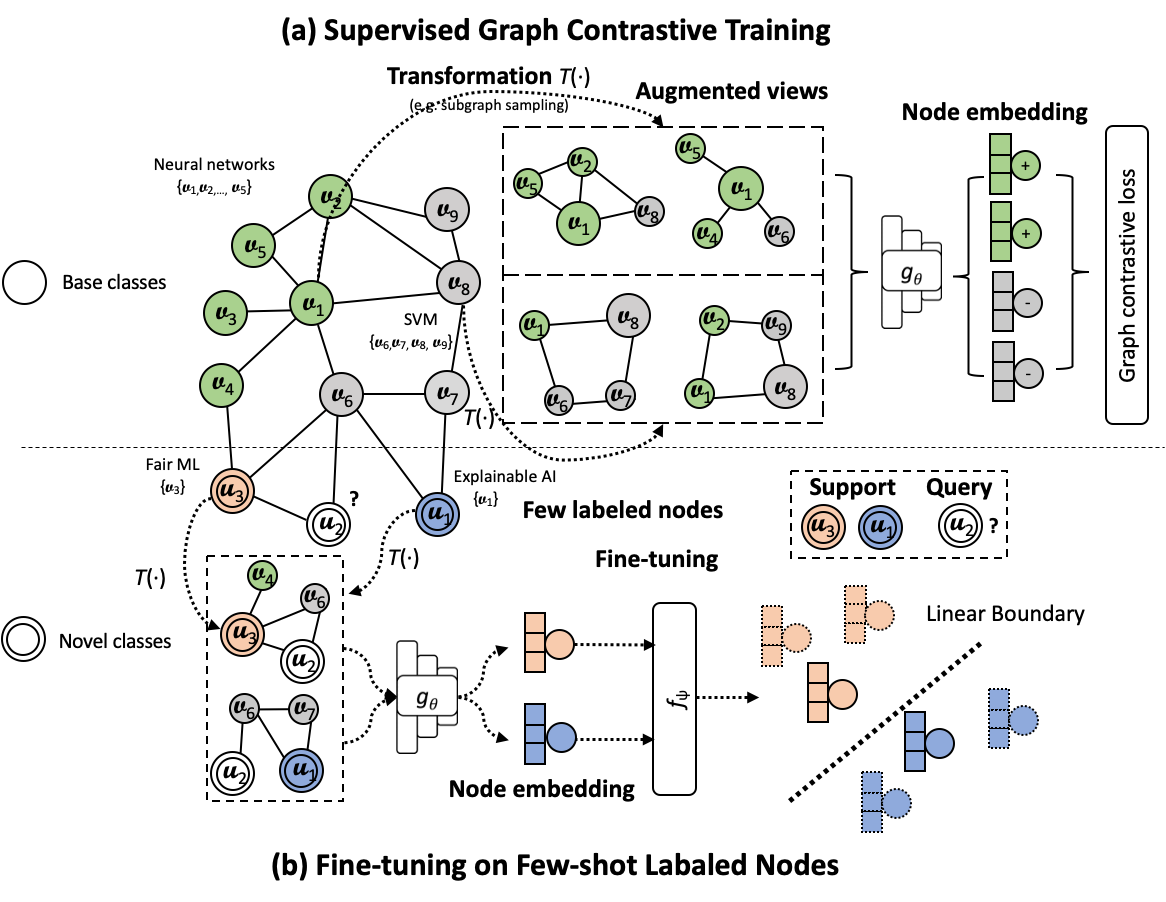}}
  \caption{The framework of TLP with supervised GCL: (a) Supervised GCL framework. (b) Fine-tuning on few-shot labeled nodes from novel classes with support and query sets. Colors indicate different classes (e.g., \textit{Neural Networks, SVM, Fair ML, Explainable AI}). Specially, white nodes mean labels of those nodes are unavailable. Labels of all nodes in base classes are available. Different types of nodes indicate if nodes are from base classes or novel classes. The counterpart of TLP with self-supervised GCL is very simliar to this, and a figure is included in Appendix \ref{app:TLP-self}.}
  \label{fig:supervised}
  \vspace{-0.0cm}
\end{figure}
\subsection{Transductive Linear Probing for Few-shot Node Classification.}
Inspired by the motivating example above, we generalize it to a new framework, \textit{Transductive Linear Probing} (TLP), for few-shot node classification. The only difference between TLP and I-GNN is that the pretraining method can be an arbitrary strategy rather than the vanilla supervised learning. It can even be self-supervised training methods that do not have any requirement on base classes. In this way, the second line of Eq. \eqref{eq:intransigent} can be generalized to:
\begin{equation}
    \theta = \arg\min_{\theta} L_{pretrain}(\mathcal{T}_{tr}^\prime;\theta),
\end{equation}
where $L_{pretrain}$ is an arbitrary loss function to pretrain the GNN encoder $g_\theta$. Then following Eq. \eqref{eq:ft}, we can exploit a linear classifier to probe the transferred embeddings of nodes from novel classes, and perform the final node classification.

In this paper, we thoroughly investigate Graph Contrastive Learning (GCL) as the pretraining strategy for TLP due to two reasons: (1) GCL \cite{hassani2020contrastive,zhu2020deep,jin2021multi,xu2021self,suresh2021adversarial,ding2023structural} is a proved effective way to learn generalizable node representations in either a supervised or self-supervised manner. By maximizing the consistency over differently transformed positive and negative examples (termed as views), GCL enforces the GNNs to be aware of the semantic and topological knowledge and injected perturbations on graphs. Trained on the global structures, GCL should be capable of addressing the piecemeal knowledge issue in meta-learning to increase the generalizability of the learned GNNs. Also, \cite{zhu2021empirical} summarizes the characteristics of GCL frameworks and empirically demonstrates the transferability of the learned representations. (2) GCL has no requirement for the base classes, which means GCL can be deployed even when the number of base classes is limited, or the nodes in base classes are unlabeled. The effectiveness of GCL highly relies on the contrastive loss function. There are two categories of contrastive loss function for graphs: (1) Supervised Contrastive Loss ($L_{SupCon}$) \cite{khosla2020supervised,akkas2022jgcl}. (2) Self-supervised Contrastive Loss: Information Noise Contrastive Estimation ($L_{InfoNCE}$) \cite{zhu2020deep,jin2021multi,mo2022simple} and Jensen-Shannon
Divergence ($L_{JSD}$) \cite{hassani2020contrastive,you2020graph}. We also consider a special GCL method, BGRL~\cite{thakoor2021large}, which does not explicitly require negative examples. The framework for TLP with an iconic supervised GCL method is provided in Fig.~\ref{fig:supervised}. From another perspective, our work is the first to focus on the extrapolation ability of GCL methods, especially under extremer few-shot settings without labels for nodes in base classes.

\section{Experimental Study}

\subsection{Experimental Settings}
We conduct systematic experiments to compare the performance of meta-learning and TLP methods (with self-supervised and supervised GCL) on the few-shot node classification task. For meta-learning, we evaluate \textbf{ProtoNet}~\cite{snell2017prototypical}, \textbf{MAML}~\cite{finn2017model}, \textbf{Meta-GNN}~\cite{zhou2019meta}, \textbf{G-Meta}~\cite{huang2020graph},
\textbf{GPN}~\cite{ding2020graph},
\textbf{AMM-GNN}~\cite{wang21AMM}, and \textbf{TENT}~\cite{wang2022task}. For TLP methods with both self-supervised and supervised forms, we evaluate  \textbf{MVGRL}~\cite{hassani2020contrastive}, \textbf{GraphCL}~\cite{you2020graph}, \textbf{GRACE}~\cite{zhu2020deep}, \textbf{MERIT}~\cite{jin2021multi}, and \textbf{SUGRL}~\cite{mo2022simple}. Moreover, \textbf{BGRL}~\cite{thakoor2021bootstrapped} and \textbf{I-GNN}~\cite{tian2020rethinking} are exclusively used for TLP methods with self-supervised GCL or supervised GCL, respectively. The detailed descriptions of these models can be found in Appendix~\ref{app_baseline}. For comprehensive studies, we benchmark those methods on six prevalent real-world graph datasets:  \texttt{CoraFull}~\cite{bojchevski2018deep},  \texttt{ogbn-arxiv}~\cite{hu2020open}, \texttt{Coauthor-CS}~\cite{shchur2018pitfalls}, \texttt{Amazon-Computer}~\cite{shchur2018pitfalls}, \texttt{Cora}~\cite{yang2016revisiting},
and \texttt{CiteSeer}~\cite{yang2016revisiting}. Specifically, each dataset is a connected graph and consists of multiple node classes for training and evaluation.
%and contains a considerable number of node classes. This is to ensure that the target test tasks consist of various classes for a more comprehensive evaluation. 
A more detailed description of those datasets is provided in Appendix \ref{app:datasets} with their statistics and class split policies in Table \ref{tab:statistics} in Appendix \ref{app:statistic}.

\subsection{Evaluation Protocol}
\label{sec:protocol}
In this section, we specify the evaluation protocol used to compare both meta-learning based methods and TLP based methods. For an attributed graph dataset $\mathcal{G} = (\mathbf{A}, \mathbf{X})$ with a divided node label space $\mathbb{C} = \{\mathbb{C}_{base}, \mathbb{C}_{novel}$ (or $\mathbb{C}_{test}$)$\}$, we split $\mathbb{C}_{base}$ into $\mathbb{C}_{train}$ and $\mathbb{C}_{dev}$ (The split policy for each datasets are listed in Table \ref{tab:statistics}). For evaluation, given a GNN encoder $g_\theta$, a classifier $f_\psi$, the validation epoch interval $V$, the number of sampled meta-tasks for evaluation $I$, the epoch patience $P$, the maximum epoch number $E$, the experiment repeated times $R$, and the $N$-way, $K$-shot, $M$-query setting specification, the final FSNC accuracy $\mathcal{A}$ and the confident interval $\mathcal{I}$ (two mainly-concerned metrics) are calculated according to Algorithm \ref{algo:protocol} given below. The default values of all those parameters are given in Table~\ref{tab:para} in Appendix \ref{app:val4protocol}.

% \section{Pseudo-Code Style Description of Evaluation Protocol}
\label{app:protocol}

\begin{algorithm}[]
\small
\caption{\textsc{Unified Evaluation Protocol for Few-shot Node Classification}}
\label{algo:protocol}
\begin{algorithmic}[1]

\REQUIRE Graph $\mathcal{G}$, $\mathbb{C}_{train}$, $\mathbb{C}_{dev}$, $\mathbb{C}_{test}$; GNN $g_\theta$, classifier $f_\psi$; parameters $V$, $I$, $P$, $E$, $R$, $N$, $K$, $M$  
\ENSURE Trained models $g_\theta$ and $f_\psi$, accuracy $\mathcal{A}$, confident interval $\mathcal{I}$.
\renewcommand{\algorithmicrequire}{\textbf{Input:}}
\renewcommand{\algorithmicensure}{\textbf{Output:}}

% \COMMENT{\slash\slash \space \texttt{Use loops for clarity. Some parts can be simplified by vectorization.}}
\slash\slash\space  \texttt{Repeat experiment for $R$ times}
\FOR{$r=1,2,\dotsc,R$}
    \STATE $p\gets1$, $t\gets1$, $s_{best}\gets0$;
    \WHILE{$t \leq E$}
        \STATE Optimize $g_\theta$ based on the specific training strategy (i.e., meta-learning and TLP); \hfill\slash\slash\space  \texttt{Training}

        \IF{$t\mod V =0$} 
        \STATE Sample $I$ meta-tasks from $\mathbb{C}_{dev}$ on $\mathcal{G}$;\hfill\slash\slash\space  \texttt{Validation}
        \STATE Calculate the obtained few-shot node classification accuracy $s$;
            \IF{$s>s_{best}$}
            \STATE $s_{best}\gets s$, $p\gets0$;
            \ELSE 
            \STATE $p\gets p+1$;
            \ENDIF
        \ENDIF
        \IF {$p=P$}
        \STATE \textbf{break}; \hfill\slash\slash\space  \texttt{Early Break}
        \ENDIF
    \ENDWHILE
    \STATE Sample $I$ meta-tasks from $\mathbb{C}_{test}$ on $\mathcal{G}$;\hfill\slash\slash\space  \texttt{Test}
    \STATE Calculate the obtained classification accuracy $s_{test}$;
    
    \STATE $s_{r}\gets s_{test}$, $r \gets r + 1$;
\ENDFOR
\STATE Calculate averaged accuracy $\mathcal{A}$ and confident interval $\mathcal{I}$ based on $\{s_1,s_2,\dotsc,s_r\}$;

%\RETURN $P$
\end{algorithmic}
\end{algorithm}

% The detailed parameter settings are provided in  Table \ref{tab:para} in Appendix \ref{app:val4protocol}.

%\subsection{Compared Models}
%\label{sec:models}

	\begin{table*}[htbp]
		\setlength\tabcolsep{4.5pt}%调列距
	\scriptsize
		\centering
		\renewcommand{\arraystretch}{1.6}
		\caption{The overall few-shot node classification results of  meta-learning methods and TLP with various GCL methods under different settings. Accuracy ($\uparrow$) and confident interval ($\downarrow$) are in $\%$. The best and second best results are \textbf{bold} and \underline{underlined}, respectively. OOM denotes out of memory.}
        \vspace{-0.0in}
		\begin{tabular}{c||c|c||c|c||c|c}
			\hline
			%\multirow{2}{*}{Model}
			Dataset&\multicolumn{2}{c||}{\texttt{CoraFull}}&\multicolumn{2}{c||}{\texttt{ogbn-arxiv}}&\multicolumn{2}{c}{\texttt{CiteSeer}}
			\\
			\hline
						Setting&\multicolumn{1}{c|}{5-way 1-shot}&\multicolumn{1}{c||}{5-way 5-shot}&\multicolumn{1}{c|}{5-way 1-shot}&\multicolumn{1}{c||}{5-way 5-shot}&\multicolumn{1}{c|}{2-way 1-shot}&\multicolumn{1}{c}{2-way 5-shot}\\
					\hline \multicolumn{7}{c}{Meta-learning}\\\hline
MAML~\cite{finn2017model}&$22.63\pm1.19$&$27.21\pm1.32$&$27.36\pm1.48$&$29.09\pm1.62$&$52.39\pm2.20$&$54.13\pm2.18$\\\hline
ProtoNet~\cite{snell2017prototypical}&$32.43\pm1.61$&$51.54\pm1.68$&$37.30\pm2.00$&$\underline{53.31\pm1.71}$&$52.51\pm2.44$&$55.69\pm2.27$\\\hline
Meta-GNN~\cite{zhou2019meta}&$55.33\pm2.43$&$70.50\pm2.02$&$27.14\pm1.94$&$31.52\pm1.71$&$\underline{56.14\pm2.62}$&$\underline{67.34\pm2.10}$\\\hline
GPN~\cite{ding2020graph}&$52.75\pm2.32$&$72.82\pm1.88$&$\underline{37.81\pm2.34}$&$50.50\pm2.13$&$53.10\pm2.39$&$63.09\pm2.50$\\\hline
AMM-GNN~\cite{wang21AMM}&$\underline{58.77\pm2.49}$&$\underline{75.61\pm1.78}$&$33.92\pm1.80$&$48.94\pm1.87$&$54.53\pm2.51$&$62.93\pm2.42$\\\hline
G-Meta~\cite{huang2020graph}&$\mathbf{60.44\pm2.48}$&$\mathbf{75.84\pm1.70}$&$31.48\pm1.70$&$47.16\pm1.73$&$55.15\pm2.68$&$64.53\pm2.35$\\\hline
TENT~\cite{wang2022task}&$55.44\pm2.08$&$70.10\pm1.73$&$\mathbf{48.26\pm1.73}$&$\mathbf{61.38\pm1.72}$&$\mathbf{62.75\pm3.23}$&$\mathbf{72.95\pm2.13}$\\
	\hline \multicolumn{7}{c}{TLP with Supervised GCL}\\\hline
I-GNN~\cite{tian2020rethinking}&$42.70\pm1.92$&$51.46\pm1.69$&$\underline{38.46\pm1.77}$&$\underline{51.46\pm1.69}$&$58.70\pm3.17$&$65.60\pm2.58$\\\hline
MVGRL~\cite{hassani2020contrastive}&$44.98\pm1.99$&$71.18\pm1.75$&OOM&OOM&$55.79\pm1.39$&$66.72\pm2.13$\\\hline
GraphCL~\cite{you2020graph}&$47.00\pm1.64$&$67.94\pm1.71$&OOM&OOM&$53.55\pm1.68$&$69.50\pm1.41$\\\hline
GRACE~\cite{zhu2020deep}&$\mathbf{65.48\pm2.45}$&$\mathbf{85.08\pm1.49}$&OOM&OOM&$61.20\pm2.39$&$\mathbf{81.76\pm1.74}$\\\hline
MERIT~\cite{jin2021multi}&$52.80\pm2.72$&$\underline{81.30\pm1.53}$&OOM&OOM&$\underline{61.25\pm2.59}$&\underline{$81.45\pm1.80$}\\\hline
SUGRL~\cite{mo2022simple}&$\underline{54.26\pm2.24}$&$77.55\pm1.95$&$\mathbf{52.13\pm2.11}$&$\mathbf{70.05\pm1.56}$&$\mathbf{65.34\pm2.55}$&$75.81\pm1.43$\\
	\hline \multicolumn{7}{c}{TLP with Self-supervised GCL}\\\hline
MVGRL~\cite{hassani2020contrastive}&$59.91\pm2.39$&$76.76\pm1.63$&OOM&OOM&$64.45\pm2.77$&$80.25\pm1.82$\\\hline
GraphCL~\cite{you2020graph}&$64.20\pm2.56$&$83.74\pm1.46$&OOM&OOM&$73.55\pm3.09$&$\mathbf{92.35\pm1.24}$\\\hline
BGRL~\cite{thakoor2021bootstrapped}&$43.83\pm2.11$&$70.44\pm1.62$&$\underline{36.76\pm1.74}$&$\underline{53.44\pm0.36}$&$54.32\pm1.63$&$70.50\pm2.11$\\\hline
GRACE~\cite{zhu2020deep}&$72.42\pm2.06$&$83.82\pm1.67$&OOM&OOM&$60.75\pm2.54$&$78.42\pm2.01$\\\hline
MERIT~\cite{jin2021multi}&$\underline{73.38\pm2.25}$&$\mathbf{87.66\pm1.43}$&OOM&OOM&$\underline{64.53\pm2.81}$&$\underline{90.32\pm1.66}$\\\hline
SUGRL~\cite{mo2022simple}&$\mathbf{77.35\pm2.20}$&$\underline{83.96\pm1.52}$&$\mathbf{60.04\pm2.11}$&$\mathbf{77.52\pm1.45}$&$\mathbf{77.34\pm2.83}$&$86.32\pm1.57$\\\hline

\end{tabular}
		\label{tab:all_result}
	\end{table*}

\subsection{Comparison}
Table~\ref{tab:all_result} presents the performance comparison of all methods on the few-shot node classification task. Specifically, we give results under four different few-shot settings to exhibit a more comprehensive comparison: 5-way 1-shot, 5-way 5-shot, 2-way 1-shot, and 2-way 5-shot. More results are given in Appendix \ref{app:moreres}. We choose the average classification accuracy and the 95\% confidence interval over $R$ repetitions as the evaluation metrics. From Table \ref{tab:all_result}, we discover the following observations:

\begin{itemize}
    \item TLP methods \textbf{consistently outperforms} meta-learning methods, which indicates the importance of transferring comprehensive node representations in FSNC tasks. In TLP methods, the model is forced to extract node-level structural information, while the meta-learning methods mainly focus on label information. As a result, TLP methods can transfer better node representations and exhibit superior performance on meta-test tasks. 
    
    \item Even \textbf{without using any label information} from base classes, TLP with self-supervised GCL methods can mostly outperform TLP with supervised GCL methods. This signifies that directly injecting supervision can potentially hinder the generalizability for TLP, which is further investigated in the following sections. 
    %Thus, we conduct more experiments on this point in the following sections. 
    %Also, this benchmark identifies that improving adaptability would be a promising direction for meta-learning based methods. 
    %The performance of TLP with self-supervised GCL consistently outperforms other baselines. This is potentially because in this strategy, the model is forced to learn more comprehensive node-level information, while the other two strategies focus more on supervised (i.e., label) information. As a result, the model can transfer better node representations for meta-test tasks.
    \item \textbf{Increasing the number of shots} $K$ (i.e., number of labeled nodes in the support set) has more significant effect on performance of both forms of TLP methods, compared with meta-learning methods. This is due to the fact that with the additional support nodes, TLP with GCL can provide more informative node representations to learn a more powerful classifier. Instead, the meta-learning methods are based on the extracted label information and thus cannot benefit from additional node-level information.
    %\item Classic few-shot learning methods (e.g., ProtoNet~\cite{snell2017prototypical} and MAML~\cite{finn2017model}) exhibit \textbf{less competitive} performance compared with other methods. This is because these methods are originally proposed in other domains, without considering the topological structures and thus lead to unsatisfactory performance on graph-structured data.
    \item Most TLP methods encounter the \textbf{OOM (out of memory) problem} when applied to the \texttt{ogbn-arxiv} dataset. This is due to the fact that the contrastive strategy in TLP methods will consume a larger memory compared with traditional supervised learning. Thus, the scalability problem is not negligible for TLP with GCL methods.
    \item BGRL~\cite{thakoor2021bootstrapped} exhibits \textbf{less competitive} performance compared with other TLP methods with self-supervised GCL. The result indicates that negative samples are important for self-supervised GCL in FSNC, which can help the model exploit node-level information. Nevertheless, without the requirement of negative samples, BGRL can parallel better to handle the OOM problem.
\end{itemize}

\subsection{Further Analysis}
\begin{figure}[htbp]
		\centering
\includegraphics[width=0.98\textwidth]{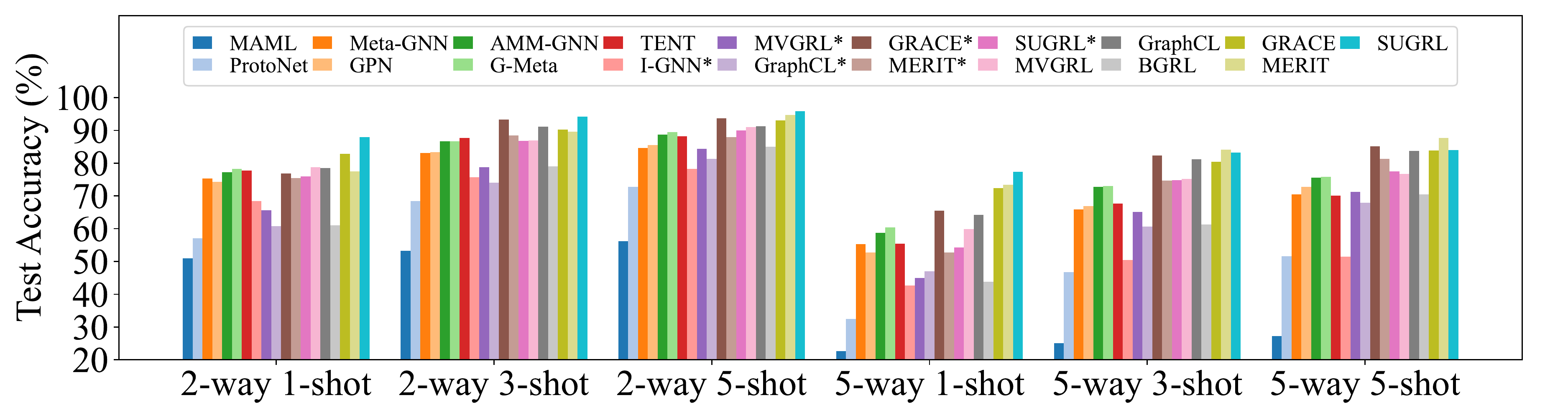}
	%\subcaptionbox{\includegraphics[width=0.48\textwidth]{img/nwaykshot.pdf}}
\caption{$N$-way $K$-shot results on \texttt{CoraFull}, meta-learning and TLP. TLP Methods with $\ast$ are based on supervised GCL methods and I-GNN.}
\label{fig:bar}
	\end{figure}

\begin{figure}[htbp]
		\centering
		\subcaptionbox{\texttt{CiteSeer}}
		{\includegraphics[width=0.48\textwidth]{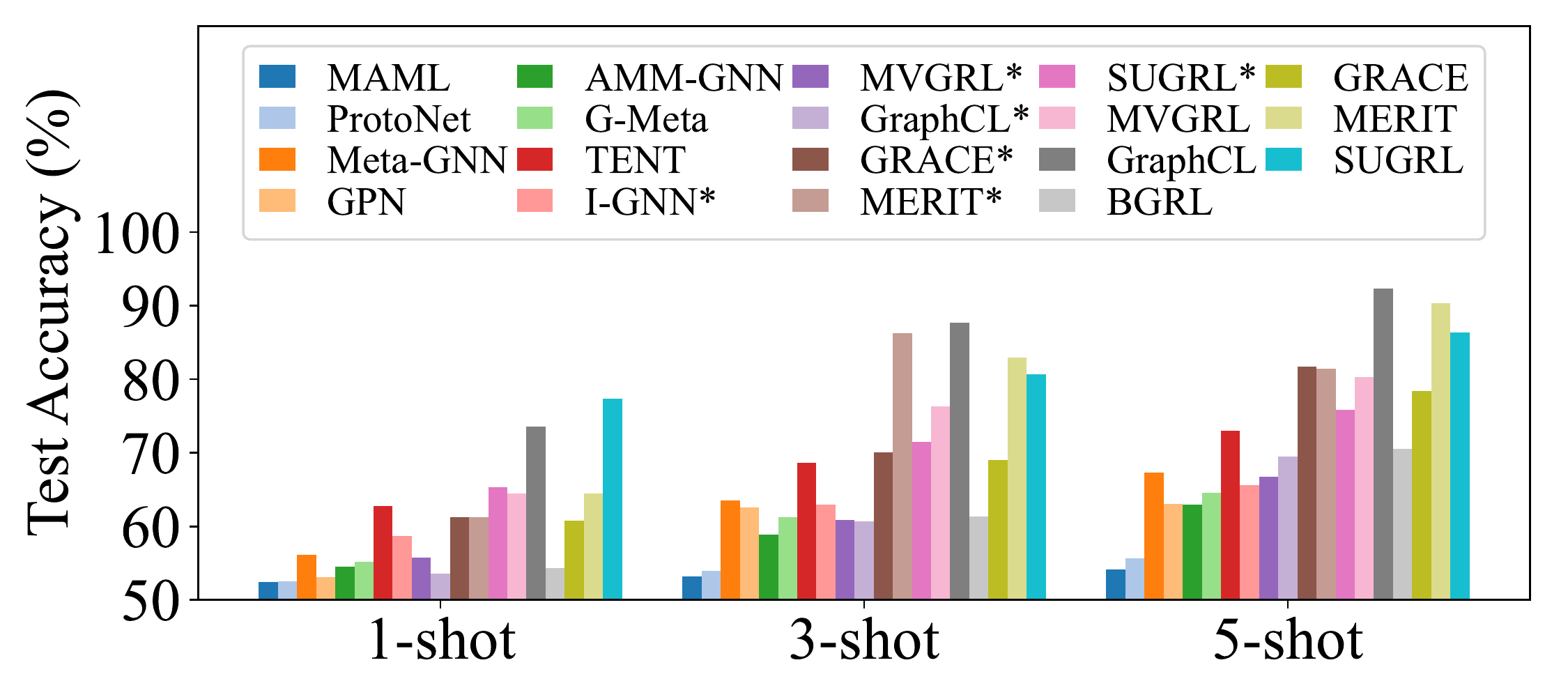}}
		\subcaptionbox{\texttt{Amazon-Computer}}
{\includegraphics[width=0.48\textwidth]{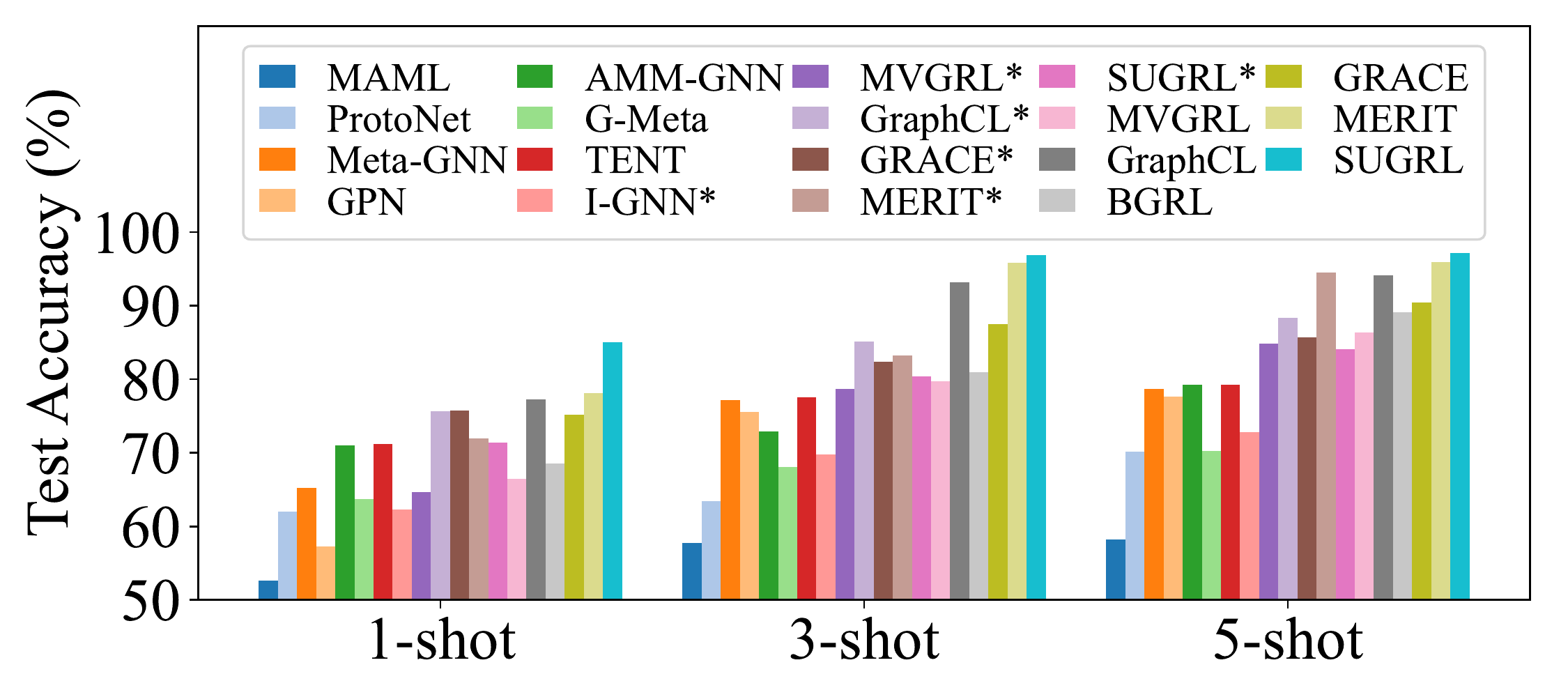}}
\caption{2-way $K$-shot results on \texttt{CiteSeer} and \texttt{Amazon-Computer}, meta-learning and two forms of TLP. TLP Methods with $\ast$ are based on supervised GCL methods and I-GNN.}
\label{fig:result_citeseer}
	\end{figure}

To explicitly compare the results between meta-learning and TLP and between two forms of TLP, we provide further results of all methods on various $N$-way $K$-shot settings in Fig.~\ref{fig:bar} and Fig.~\ref{fig:result_citeseer}. From the results, we can obtain the following observations:
\begin{itemize}
    \item  \textcolor{black}{When a larger values of $N$ is presented, the performance drop is \textbf{less significant} on TLP based methods compared to meta-learning based methods. The performance of all methods degrades as $N$ increases (i.e., more classes in each meta-task). With a larger $N$, the variety of classes in each meta-task can result in a more complex class distribution and thus increase the classification difficulties. Nevertheless, the performance drop is less significant on TLP with both forms of GCL methods. This is because the utilized GCL methods focus more on node-level structural patterns, which incorporate more potentially useful information for classification. As a result, TLP is more capable of alleviating the problem of difficult classification caused by a larger $N$.}
    
    % The performance of all methods \textbf{significantly degrades} when a larger value of $N$ is presented (i.e., more classes in each meta-task). The main reason is that with a larger $N$, the variety of classes in each meta-task can result in a more complex class distribution and thus increase the classification difficulties. Nevertheless, the performance drop is less significant on TLP with both forms of GCL methods. This is because the utilized GCL methods focus more on node-level structural patterns, which incorporate more potentially useful information for classification. As a result, these methods are more capable of alleviating the problem of difficult classification caused by a larger $N$.
    
    \item 
    %Among the three 2-way datasets, the performance improvement of GCL over meta-learning on \texttt{Amazon-Computer} is generally more impressive than the other two datasets. 
    As shown in Fig.~\ref{fig:result_citeseer}, the \textbf{performance improvement} of TLP with self-supervised GCL methods over meta-learning methods on \texttt{CiteSeer} is generally more impressive than other datasets.
    The main reason is that  \texttt{CiteSeer} bears a significantly smaller class set (2/2/2 classes for $\mathbb{C}_{train}$/$\mathbb{C}_{dev}$/$\mathbb{C}_{test}$). In consequence, the meta-learning methods cannot effectively leverage the supervision information during training. Nevertheless, TLP with self-supervised GCL can extract useful structural information for better generalization performance.
    %average node degree value. In consequence, 
    %there exists richer node-level structural information on the graph, which can be further extracted and 
    %the node-level structural information on the graph can be crucial for classification, which is extracted and utilized by TLP methods via the constrative strategy for better performance. 
\end{itemize}

\subsection{Effect of Supervision Information in Base Classes}
\begin{figure}[htbp]
		\centering

		\subcaptionbox{TLP with self-supervised and supervised ($*$) GCL}
{\includegraphics[width=0.48\textwidth]{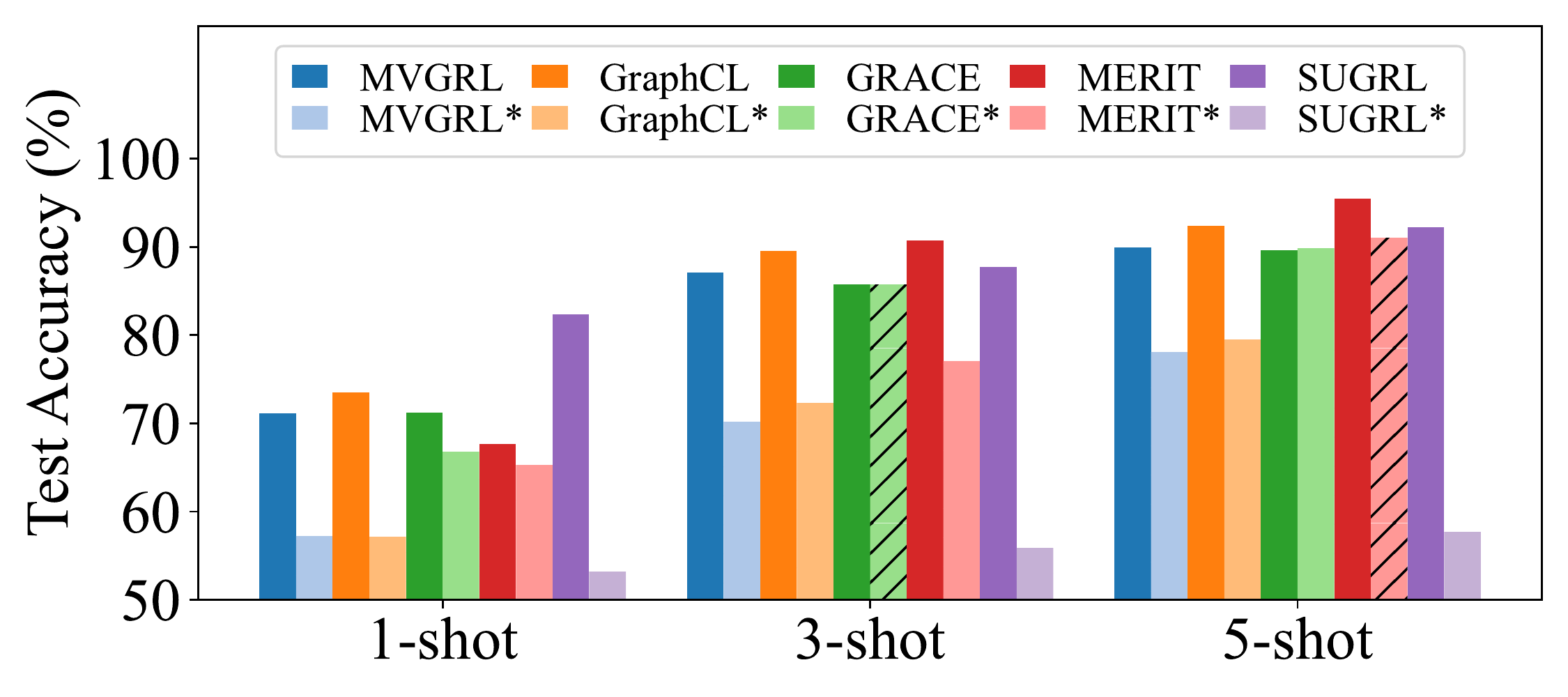}}
	\subcaptionbox{Results on 5-shot setting with different $\lambda$}
{\includegraphics[width=0.48\textwidth]{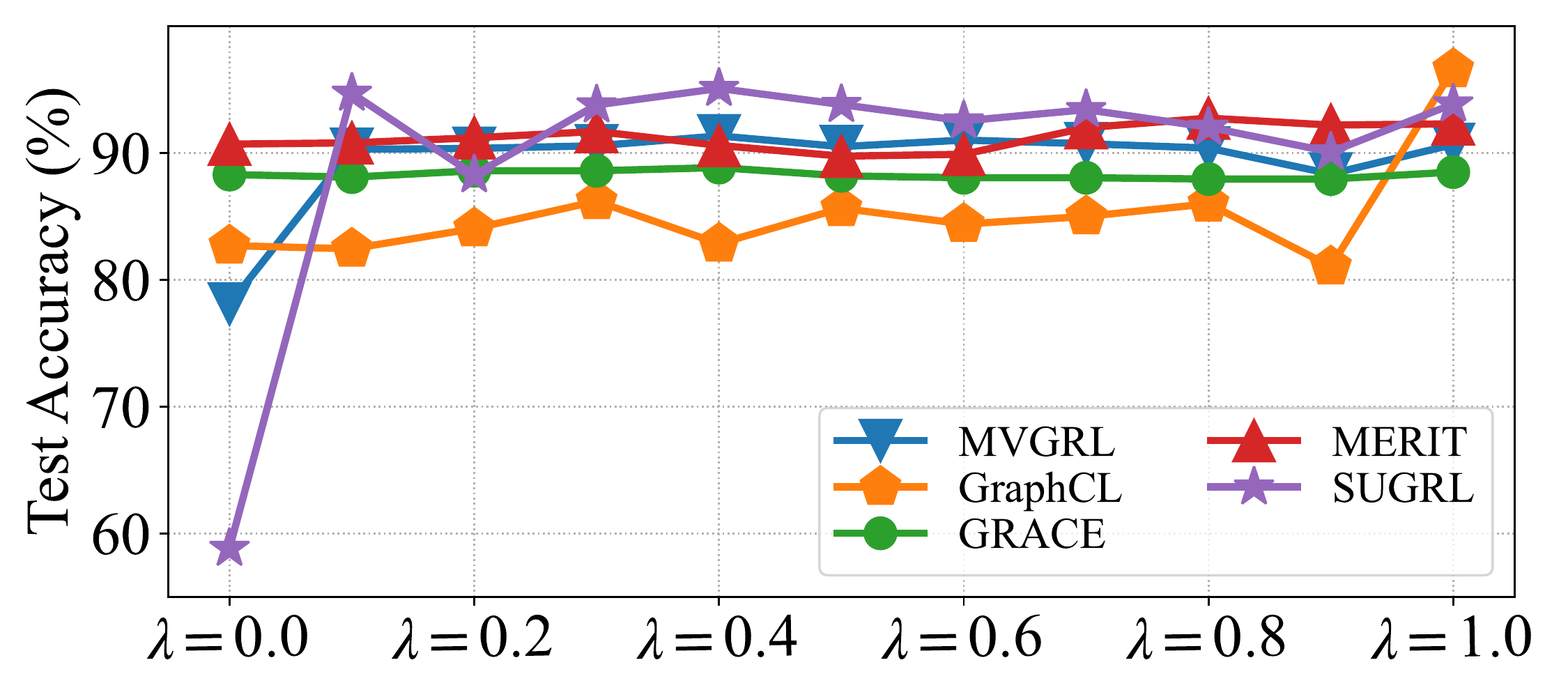}}

\caption{Results on dataset \texttt{Cora} (2-way)}
\label{fig:lambda}
	\end{figure}

In this section, we further investigate the effectiveness of the supervised information in TLP with supervised GCL methods. Specifically, we leverage a combined loss \textcolor{black}{$L_{JointCon}=\lambda L_{SelfCon} + (1-\lambda)L_{SupCon}$}, where $L_{SelfCon}$ indicates a self-supervised GCL loss, either $L_{JSD}$ or $L_{InfoNCE}$ according to the models, and $L_{JointCon}$ is a mixture of supervised GCL loss and self-supervised GCL loss. In this way, we can gradually adjust the value of $\lambda$ to inject different levels of supervision 
signals into GCL and then observe the performance fluctuation. Note that due to the unstable training curve brought by the joint loss $L_{JointCon}$, we increase the epoch patience number from $P$ to $2P$ to ensure convergence. The results on \texttt{Cora} dataset (we observe similar results on other datasets) with different values of $\lambda$ are provided in Fig.~\ref{fig:lambda}. From the results, we can obtain the following observations: 
\begin{itemize}
    \item In general, the classification performance \textbf{increases with a larger value of $\lambda$}. In other words, directly injecting supervision information into GCL for TLP will usually reduce the performance on few-shot node classification tasks. Nevertheless, carefully injecting supervision information can slightly increase the accuracy by choosing a suitable value of $\lambda$. On the other hand, the results also verify that the TLP framework can still achieve considerable performance without any explicit restrictions for base classes.
    %On the other hand, this experiment shows that, if exclusively based on self-supervised GCL, our proposed TLP paradigm does not require any assumption for base classes and can still perform well. 
    \item Even with \textbf{a relatively small value of $\lambda$} (e.g., 0.1), the performance improvement over TLP with totally supervised GCL (i.e., $\lambda=0.0$) is still significant. That being said, the contrastive strategy that leverages graph structures can provide better performance by providing comprehensive node representations.
\end{itemize}
	
\subsection{Evaluating Learned Node Representations on Novel Classes}
In this section, we further validate the quality of the learned node representations from different training strategies. Particularly, we leverage two prevalent clustering evaluation metrics: \textit{normalized mutual information} (NMI)  and \textit{adjusted random index} (ARI), on learned node representations clustered based on K-Means. We evaluate the representations learned from two datasets \texttt{CoraFull} and \texttt{CiteSeer} for a fair comparison. The results are presented in Table~\ref{tab:nmi_result} in  Appendix~\ref{app:evaluation} . Based on the results, we can obtain the following observations:
\begin{itemize}
    \item The meta-learning methods typically exhibit \textbf{inferior NMI and ARI scores} compared with both forms of TLP. This is because meta-learning methods are dedicated for extracting supervision information from node samples and thus cannot fully utilize node-level structural information.
    \item In general, TLP with self-supervised GCL methods can result in \textbf{larger values of both NMI and ARI scores} than TLP with supervised GCL. This is due to the fact that the self-supervised GCL model focuses more on extracting structural information without the interruption of label information. As a result, the learned node representations are more comprehensive and thus exhibit superior clustering performance.
    \item \textbf{The difference of NMI and ARI score}s between meta-learning and TLP is more significant on \texttt{CiteSeer} than \texttt{CoraFull}. This phenomenon potentially results from the fact that \texttt{CiteSeer} consists of fundamentally fewer classes than \texttt{CoraFull}. In consequence, for \texttt{CiteSeer}, the meta-learning methods will largely rely on label information instead of node-level structural information for classification.

\end{itemize}

\begin{figure}[htbp]
		\centering
	\subcaptionbox{GraphCL$*$}{
\includegraphics[width=0.23\textwidth]{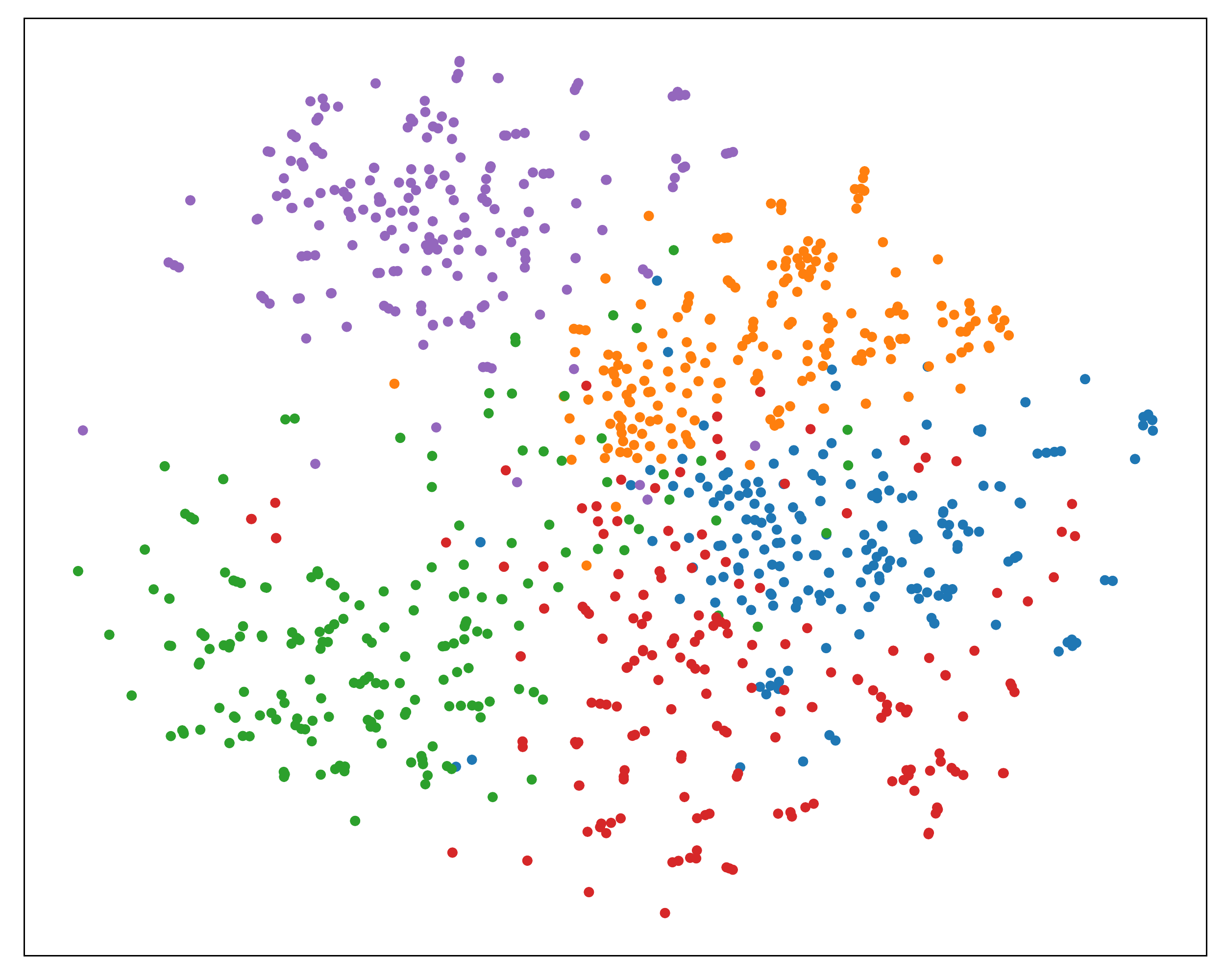}}
			\subcaptionbox{GraphCL}{
\includegraphics[width=0.23\textwidth]{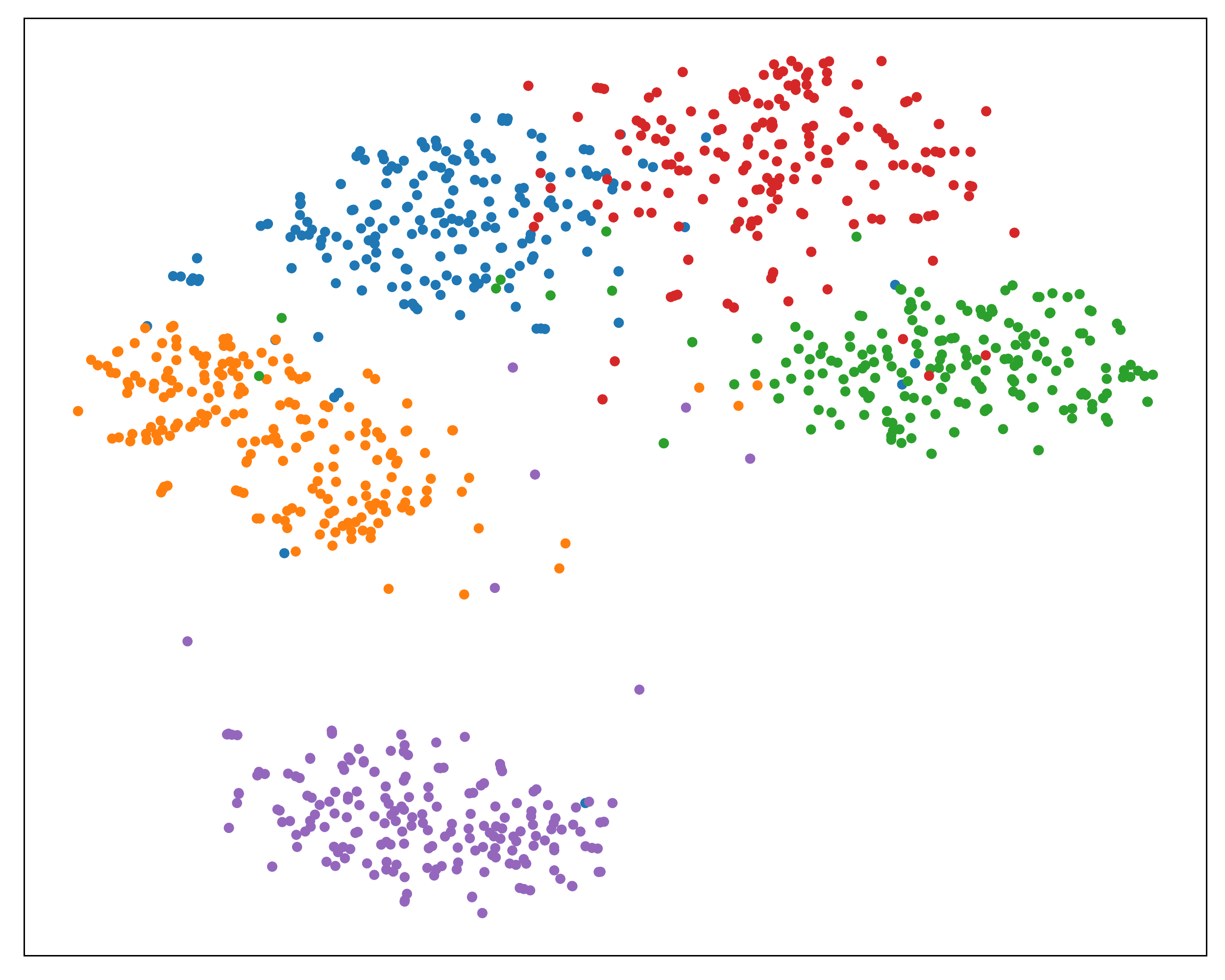}}
	\subcaptionbox{SUGRL$*$}{
\includegraphics[width=0.23\textwidth]{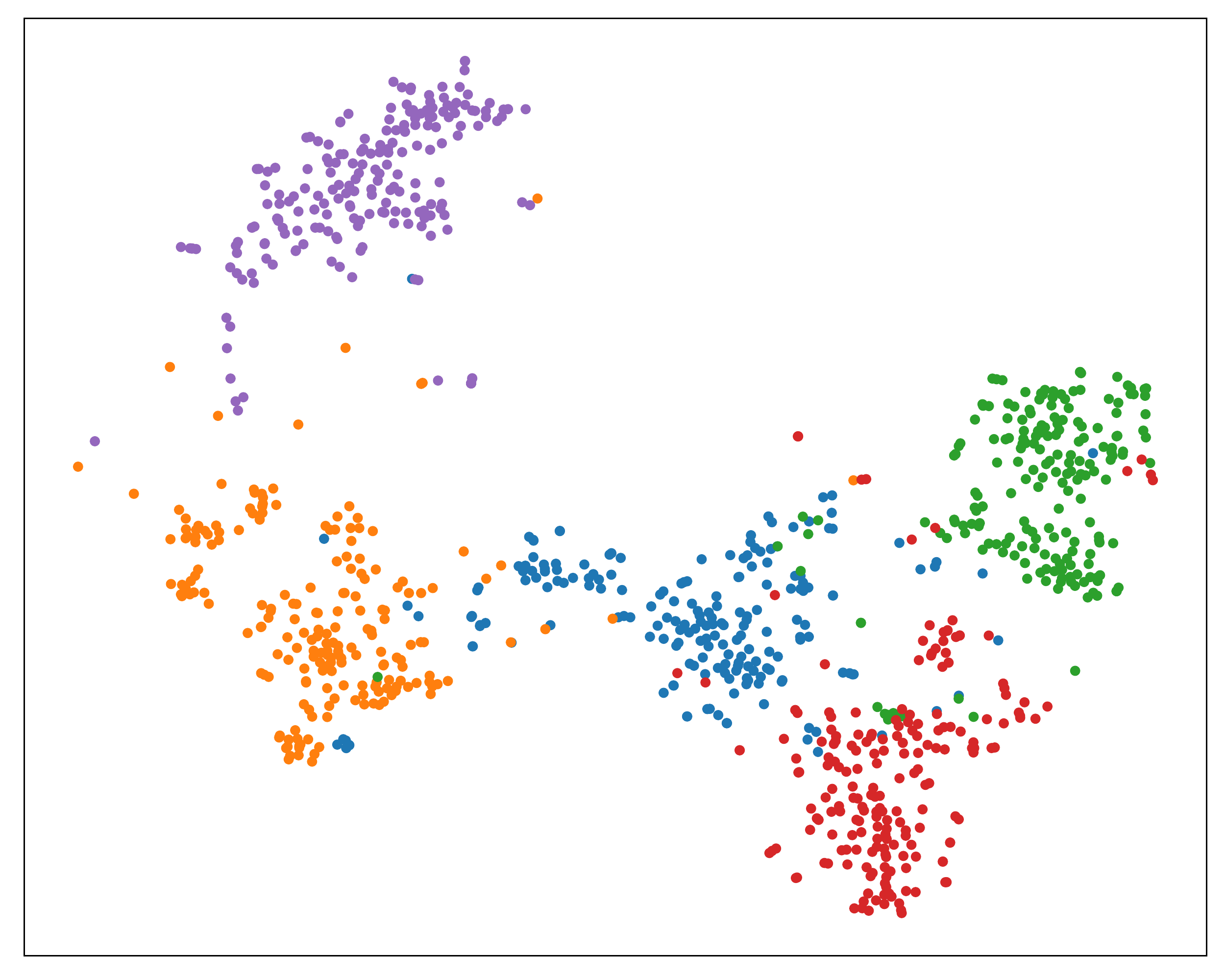}}
		\subcaptionbox{SUGRL}{
\includegraphics[width=0.23\textwidth]{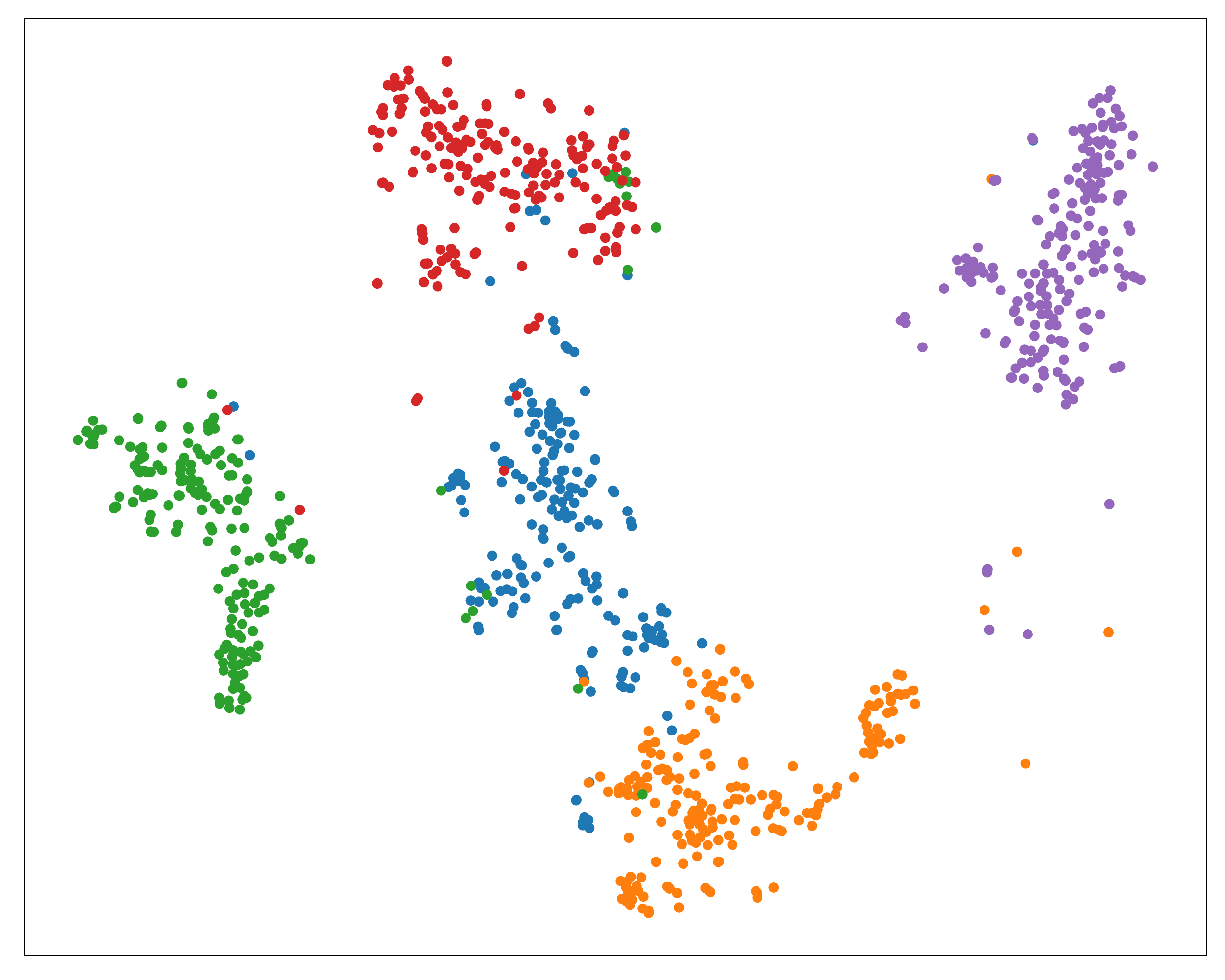}}

	\subcaptionbox{Meta-GNN}{
\includegraphics[width=0.23\textwidth]{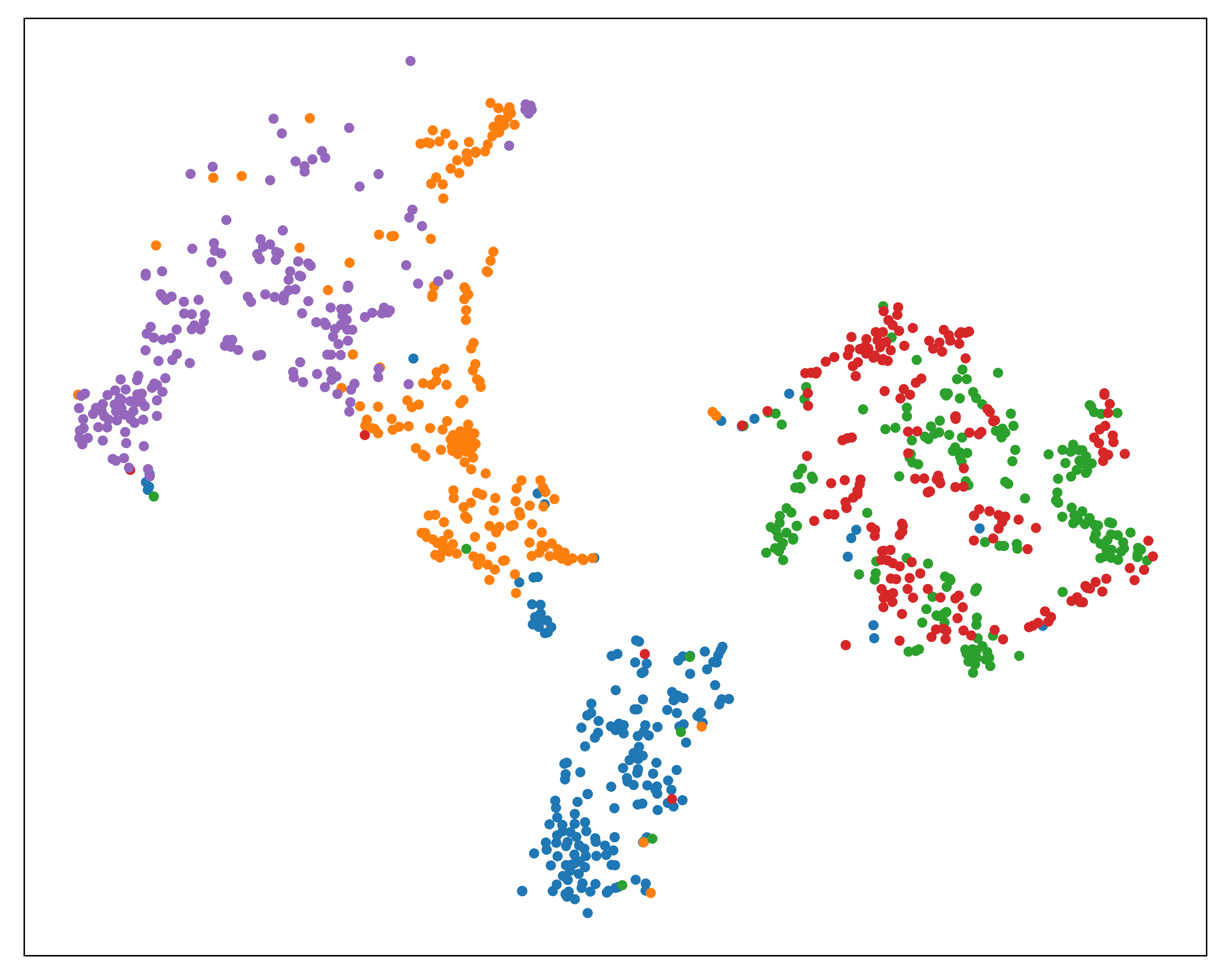}}
	\subcaptionbox{TENT}{
\includegraphics[width=0.23\textwidth]{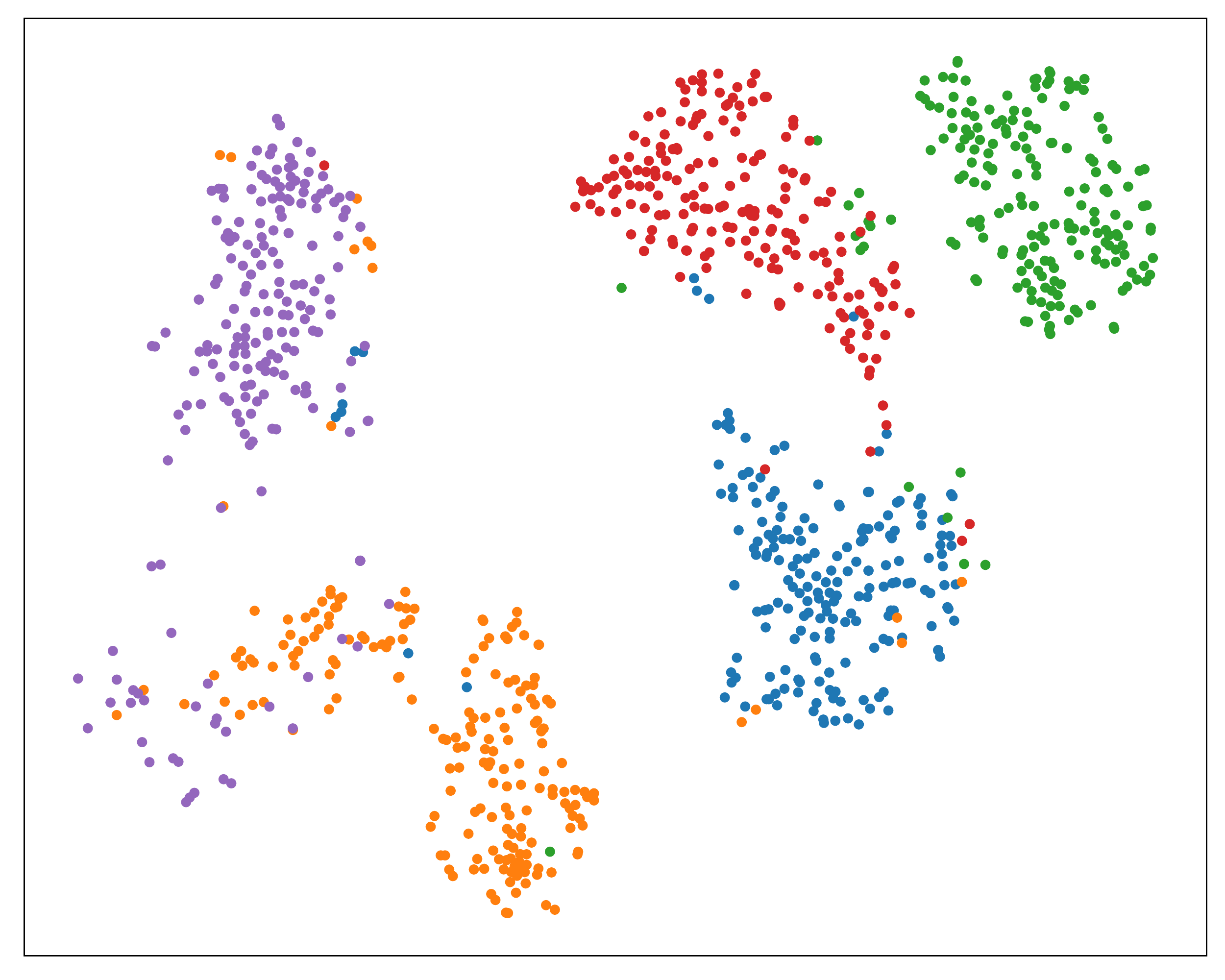}}
	\subcaptionbox{Meta-GNN}{
\includegraphics[width=0.23\textwidth]{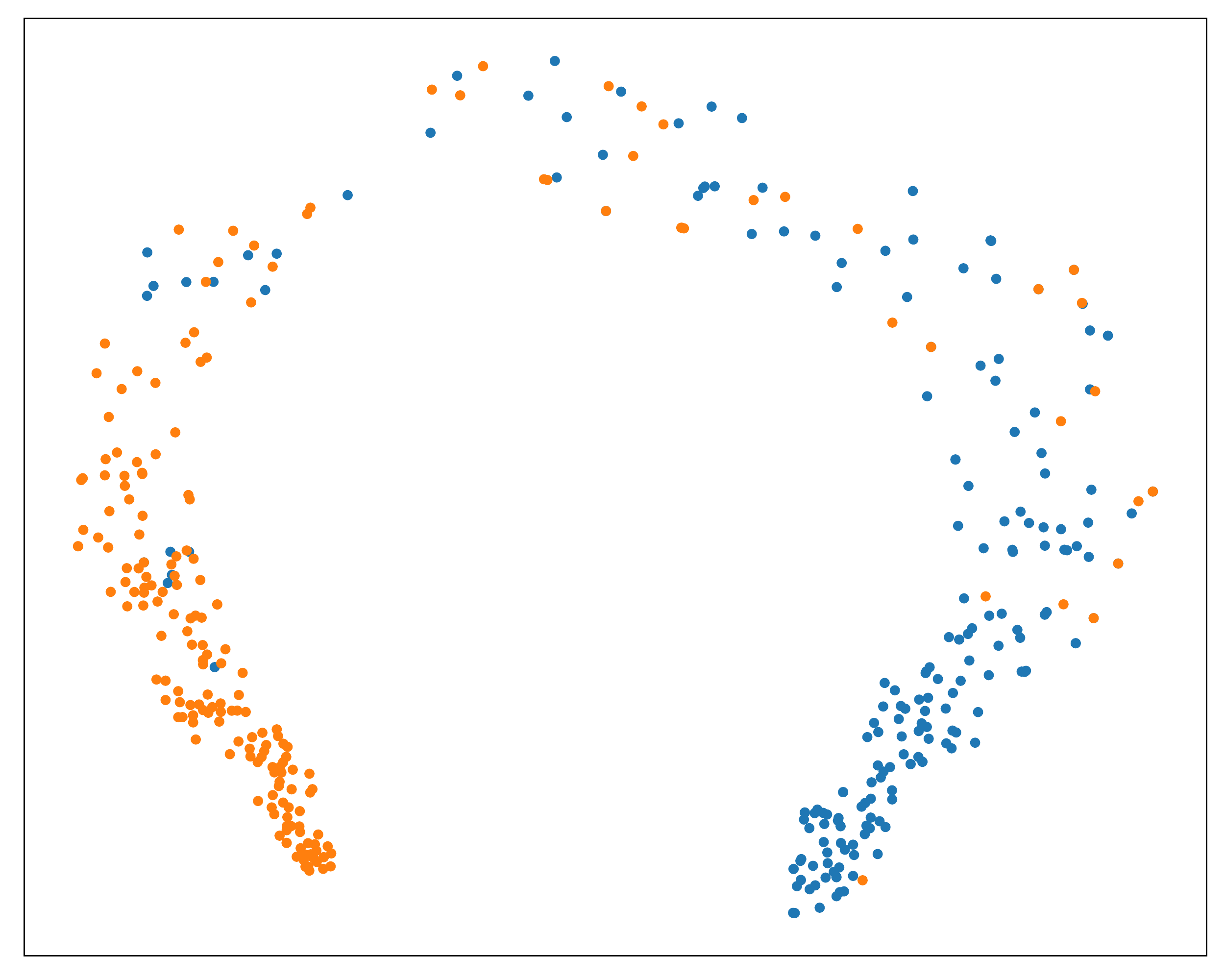}}
	\subcaptionbox{TENT}{
\includegraphics[width=0.23\textwidth]{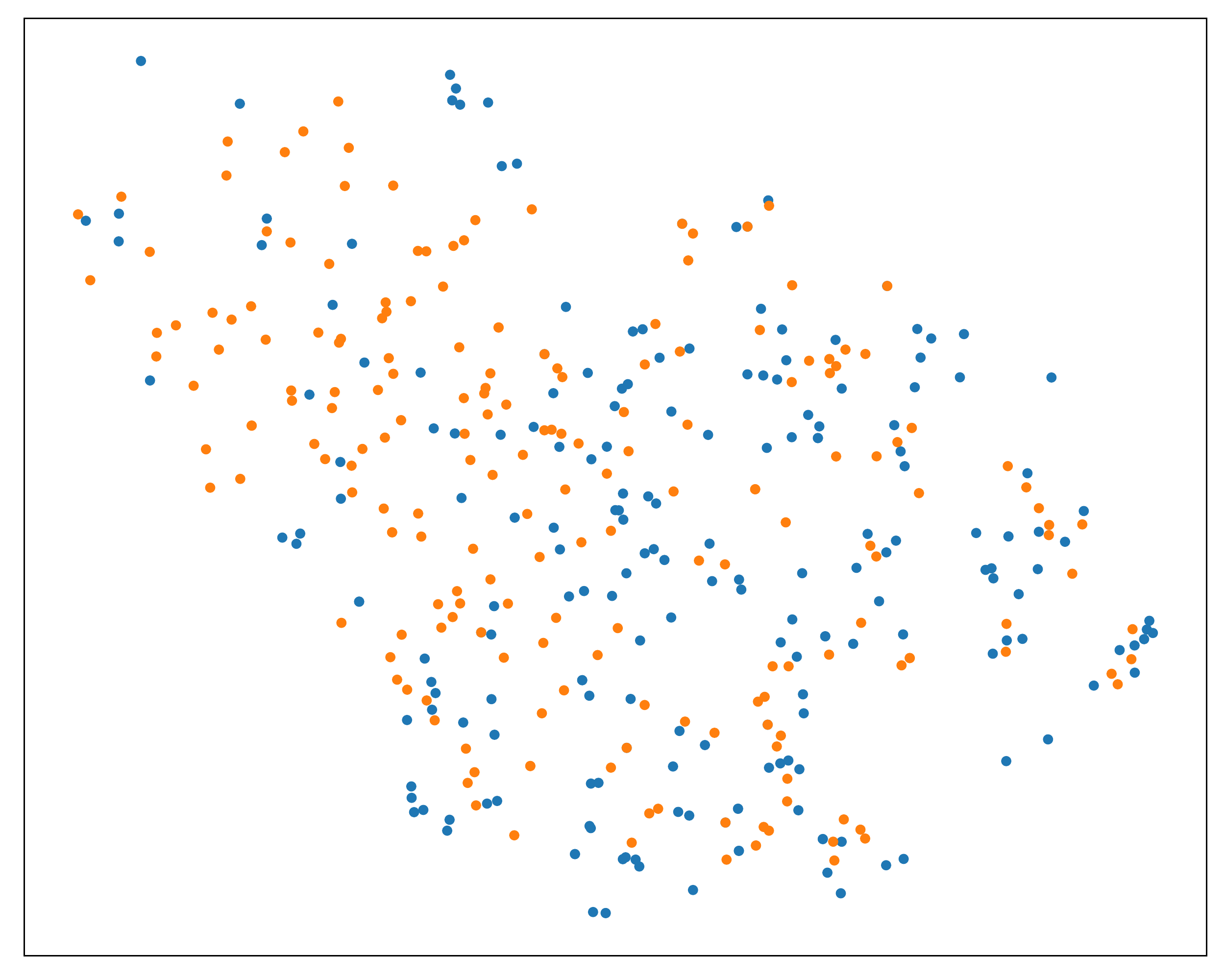}}

\caption{The t-SNE visualization results. Fig. (a)-(f) are for dataset \texttt{CoraFull} (5-way). Fig. (g)-(h) are for dataset \texttt{CiteSeer} (2-way). TLP methods with $*$ are based on supervised GCL methods.}
\vspace{-0.cm}
\label{fig:tsne}
	\end{figure}

\subsection{Visualization}
To provide an explicit comparison of different baselines, we visualize the learned node representations from \texttt{CoraFull} and \texttt{CiteSeer} via the t-SNE algorithm, where colors denote different classes. It is noteworthy that for clarity, we randomly select five classes from $\mathbb{C}_{test}$ for the visualization. The results are provided in Fig.~\ref{fig:tsne} (more results are included in Fig.~\ref{fig:tsne_extra}). Specifically, we discover that:

\begin{itemize}
    \item TLP with \textbf{self-supervised GCL} generally outperforms TLP with supervised GCL. This is because without learning label information, TLP with self-supervised GCL can concentrate on node representation patterns, which are easier to transfer to target unseen novel classes.
    \item The learned node representations are \textbf{less discriminative} for meta-learning on \texttt{CiteSeer} compared with \texttt{CoraFull}. This is because \texttt{CiteSeer} contains fewer classes, which means the node representations learned by meta-learning methods will be less informative, since they are only required to classify nodes from a small class set.
\end{itemize}

\section{Conclusion, Limitations, and Outlook}

In this paper, we propose TLP as an alternative framework to meta-learning for FSNC tasks. First, we provide a motivating example, a vanilla intransigent GNN model, to validate our postulation that a generalizable GNN encoder is the key to FSNC tasks. Then, we provide a formal definition for TLP, which transfers node embeddings from GCL pretraining to the prevailing meta-learning paradigm. We conduct comprehensive experiments and compare various meta-learning based and TLP-based methods under the same protocol. Our rigorous empirical study reveals several interesting findings on the strengths and weaknesses of the two approaches and identifies that adaptability and scalability are the promising directions for meta-learning based and TLP-based methods, respectively. 

However, due to limited space, several limitations of our work need to be acknowledged. 
\begin{itemize}

\item\textbf{Limited design considerations.} Even though an exhaustive survey on FSNC or GCL is out of the scope of this work, we do not provide a more fine-grained comparison on model details, such as different GNN encoders or various transformations during GCL pretraining. \textcolor{black}{Also, we only consider methods applied on a single graph, which currently are the mainstream of research on FSNC. There are more recent works (e.g., \cite{wang2022graph}) studying FSNC across multiple graphs}.
\item\textbf{Lack of theoretical justifications.} Our findings are based on empirical studies, which cannot disclose the underlying mathematical mechanisms of those methods, such as the performance guarantee by transferring node embeddings from different GCL methods. 
\end{itemize}

How to address these limitations is saved as future work. Note that based on the experiments here, the observations drawn are not conclusive termination. We only cover existing methods in this work and hope this work to be inspiring for developing meta-learning based FSNC methods that can outperform TLP based methods, or better ways to utilize labels in TLP methods. In broader terms, this work lies at the confluence of graph few-shot learning and graph contrastive learning. We hope this work can facilitate the sharing of insights for both communities. On the one hand, we hope our work provides a necessary yardstick to measure progress across the FSNC field. On the other hand, our work should have exhibited several practical guidelines for future research in both vigorous fields. For example, the meta-learning community can get inspired by GCL to learn more transferable graph patterns. Also, few-shot TLP can serve as a new metric to evaluate the extrapolation ability of GCL methods.

% \section*{Author Contributions}
% Authors of accepted papers are \emph{encouraged} to include a statement that declares the individual contribution of every author, especially when there are co-authors that made equal contributions to the research.
% You may adopt the \href{https://credit.niso.org/}{Contributor Roles Taxonomy (CRediT)} methodology for attributing contributions.
% Do not include this section in the version for blind review.
% This section does not count towards the page limit.

\section*{Acknowledgements}
This work is supported by the National Science Foundation under grants IIS-2006844, IIS-2144209, IIS-2223769, IIS-2229461, CNS-2154962, and BCS-2228534, the Army Research Office (ARO)
W911NF2110030, the Office of Naval Research N00014-21-1-4002, the JP Morgan Chase Faculty Research Award, the Cisco Faculty Research Award, and the Jefferson Lab subcontract JSA-22-D0311. 

% The \LaTeX{} template of LoG 2022 is heavily borrowed from NeurIPS 2022.

% Do not include acknowledgements in the version for blind review.
% If a paper is accepted, please place such acknowledgements in an unnumbered section at the end of the paper, immediately before the references.
% The acknowledgements do not count towards the page limit.

\newpage
% For natbib users:
\bibliographystyle{unsrtnat}
\bibliography{reference}

\begin{thebibliography}{58}
\providecommand{\natexlab}[1]{#1}
\providecommand{\url}[1]{\texttt{#1}}
\expandafter\ifx\csname urlstyle\endcsname\relax
  \providecommand{\doi}[1]{doi: #1}\else
  \providecommand{\doi}{doi: \begingroup \urlstyle{rm}\Url}\fi

\bibitem[Kipf and Welling(2017)]{Kipf:2017tc}
Thomas~N. Kipf and Max Welling.
\newblock {Semi-Supervised Classification with Graph Convolutional Networks}.
\newblock In \emph{ICLR}, 2017.

\bibitem[Veli{\v c}kovi{\'c} et~al.(2018)Veli{\v c}kovi{\'c}, Cucurull,
  Casanova, Romero, Li{\`o}, and Bengio]{Velickovic:2018we}
Petar Veli{\v c}kovi{\'c}, Guillem Cucurull, Arantxa Casanova, Adriana Romero,
  Pietro Li{\`o}, and Yoshua Bengio.
\newblock {Graph Attention Networks}.
\newblock In \emph{ICLR}, 2018.

\bibitem[Hamilton et~al.(2017)Hamilton, Ying, and Leskovec]{Hamilton:2017tp}
William~L. Hamilton, Zhitao Ying, and Jure Leskovec.
\newblock {Inductive Representation Learning on Large Graphs}.
\newblock In \emph{NeurIPS}, pages 1024--1034, 2017.

\bibitem[Xu et~al.(2019)Xu, Hu, Leskovec, and Jegelka]{xu2018powerful}
Keyulu Xu, Weihua Hu, Jure Leskovec, and Stefanie Jegelka.
\newblock How powerful are graph neural networks?
\newblock In \emph{Proceedings of the 2019 International Conference on Learning
  Representations}, 2019.

\bibitem[Ding et~al.(2022{\natexlab{a}})Ding, Wang, Caverlee, and
  Liu]{ding2022meta}
Kaize Ding, Jianling Wang, James Caverlee, and Huan Liu.
\newblock Meta propagation networks for graph few-shot semi-supervised
  learning.
\newblock AAAI, 2022{\natexlab{a}}.

\bibitem[Ding et~al.(2022{\natexlab{b}})Ding, Xu, Tong, and Liu]{ding2022data}
Kaize Ding, Zhe Xu, Hanghang Tong, and Huan Liu.
\newblock Data augmentation for deep graph learning: A survey.
\newblock \emph{ACM SIGKDD Explorations Newsletter}, 2022{\natexlab{b}}.

\bibitem[Zhang et~al.(2018)Zhang, Zhou, Huang, and Wei]{zhang2018few}
Shengzhong Zhang, Ziang Zhou, Zengfeng Huang, and Zhongyu Wei.
\newblock Few-shot classification on graphs with structural regularized gcns.
\newblock In \emph{Proceedings of the 32nd AAAI Conference on Artificial
  Intelligence}, 2018.

\bibitem[Ding et~al.(2020)Ding, Wang, Li, Shu, Liu, and Liu]{ding2020graph}
Kaize Ding, Jianling Wang, Jundong Li, Kai Shu, Chenghao Liu, and Huan Liu.
\newblock Graph prototypical networks for few-shot learning on attributed
  networks.
\newblock In \emph{CIKM}, 2020.

\bibitem[Wang et~al.(2020{\natexlab{a}})Wang, Luo, Ding, Zhang, Li, and
  Zheng]{wang21AMM}
Ning Wang, Minnan Luo, Kaize Ding, Lingling Zhang, Jundong Li, and Qinghua
  Zheng.
\newblock Graph few-shot learning with attribute matching.
\newblock In \emph{Proceedings of the 29th ACM International Conference on
  Information and Knowledge Management}, 2020{\natexlab{a}}.

\bibitem[Huang and Zitnik(2020)]{huang2020graph}
Kexin Huang and Marinka Zitnik.
\newblock Graph meta learning via local subgraphs.
\newblock In \emph{NeurIPS}, 2020.

\bibitem[Lan et~al.(2020)Lan, Wang, Du, Song, Tao, and Guan]{lan2020node}
Lin Lan, Pinghui Wang, Xuefeng Du, Kaikai Song, Jing Tao, and Xiaohong Guan.
\newblock Node classification on graphs with few-shot novel labels via meta
  transformed network embedding.
\newblock \emph{Advances in Neural Information Processing Systems},
  33:\penalty0 16520--16531, 2020.

\bibitem[Wang et~al.(2022{\natexlab{a}})Wang, Ding, Zhang, Chen, and
  Li]{wang2022task}
Song Wang, Kaize Ding, Chuxu Zhang, Chen Chen, and Jundong Li.
\newblock Task-adaptive few-shot node classification.
\newblock \emph{arXiv preprint arXiv:2206.11972}, 2022{\natexlab{a}}.

\bibitem[Zhang et~al.(2022)Zhang, Ding, Li, Zhang, Ye, Chawla, and
  Liu]{zhang2022few}
Chuxu Zhang, Kaize Ding, Jundong Li, Xiangliang Zhang, Yanfang Ye, Nitesh~V
  Chawla, and Huan Liu.
\newblock Few-shot learning on graphs: A survey.
\newblock \emph{arXiv preprint arXiv:2203.09308}, 2022.

\bibitem[Tan et~al.()Tan, Ding, Guo, and Liu]{tansupervised}
Zhen Tan, Kaize Ding, Ruocheng Guo, and Huan Liu.
\newblock Supervised graph contrastive learning for few-shot node
  classification.

\bibitem[Dhillon et~al.(2019)Dhillon, Chaudhari, Ravichandran, and
  Soatto]{dhillon2019baseline}
Guneet~Singh Dhillon, Pratik Chaudhari, Avinash Ravichandran, and Stefano
  Soatto.
\newblock A baseline for few-shot image classification.
\newblock In \emph{International Conference on Learning Representations}, 2019.

\bibitem[Tian et~al.(2020)Tian, Wang, Krishnan, Tenenbaum, and
  Isola]{tian2020rethinking}
Yonglong Tian, Yue Wang, Dilip Krishnan, Joshua~B Tenenbaum, and Phillip Isola.
\newblock Rethinking few-shot image classification: a good embedding is all you
  need?
\newblock \emph{Proceedings of the 16th European Conference on Computer
  Vision}, 2020.

\bibitem[Hassani and Khasahmadi(2020)]{hassani2020contrastive}
Kaveh Hassani and Amir~Hosein Khasahmadi.
\newblock Contrastive multi-view representation learning on graphs.
\newblock In \emph{International Conference on Machine Learning}, pages
  4116--4126. PMLR, 2020.

\bibitem[You et~al.(2020)You, Chen, Sui, Chen, Wang, and Shen]{you2020graph}
Yuning You, Tianlong Chen, Yongduo Sui, Ting Chen, Zhangyang Wang, and Yang
  Shen.
\newblock Graph contrastive learning with augmentations.
\newblock \emph{NeurIPS}, 2020.

\bibitem[Zhu et~al.(2020)Zhu, Xu, Yu, Liu, Wu, and Wang]{zhu2020deep}
Yanqiao Zhu, Yichen Xu, Feng Yu, Qiang Liu, Shu Wu, and Liang Wang.
\newblock Deep graph contrastive representation learning.
\newblock \emph{arXiv preprint arXiv:2006.04131}, 2020.

\bibitem[Jin et~al.(2021)Jin, Zheng, Li, Gong, Zhou, and Pan]{jin2021multi}
Ming Jin, Yizhen Zheng, Yuan-Fang Li, Chen Gong, Chuan Zhou, and Shirui Pan.
\newblock Multi-scale contrastive siamese networks for self-supervised graph
  representation learning.
\newblock In \emph{International Joint Conference on Artificial Intelligence
  2021}, pages 1477--1483. Association for the Advancement of Artificial
  Intelligence (AAAI), 2021.

\bibitem[Thakoor et~al.(2021{\natexlab{a}})Thakoor, Tallec, Azar, Azabou, Dyer,
  Munos, Veli{\v{c}}kovi{\'c}, and Valko]{thakoor2021large}
Shantanu Thakoor, Corentin Tallec, Mohammad~Gheshlaghi Azar, Mehdi Azabou,
  Eva~L Dyer, Remi Munos, Petar Veli{\v{c}}kovi{\'c}, and Michal Valko.
\newblock Large-scale representation learning on graphs via bootstrapping.
\newblock \emph{arXiv preprint arXiv:2102.06514}, 2021{\natexlab{a}}.

\bibitem[Mo et~al.(2022)Mo, Peng, Xu, Shi, and Zhu]{mo2022simple}
Yujie Mo, Liang Peng, Jie Xu, Xiaoshuang Shi, and Xiaofeng Zhu.
\newblock Simple unsupervised graph representation learning.
\newblock AAAI, 2022.

\bibitem[Ding et~al.(2023)Ding, Wang, Yang, and Liu]{ding2023structural}
Kaize Ding, Yancheng Wang, Yingzhen Yang, and Huan Liu.
\newblock Eliciting structural and semantic global knowledge in unsupervised
  graph contrastive learning.
\newblock 2023.

\bibitem[Chen et~al.(2020)Chen, Kornblith, Norouzi, and Hinton]{chen2020simple}
Ting Chen, Simon Kornblith, Mohammad Norouzi, and Geoffrey Hinton.
\newblock A simple framework for contrastive learning of visual
  representations.
\newblock In \emph{International conference on machine learning}, pages
  1597--1607. PMLR, 2020.

\bibitem[Mishra et~al.(2018)Mishra, Rohaninejad, Chen, and
  Abbeel]{mishra2018simple}
Nikhil Mishra, Mostafa Rohaninejad, Xi~Chen, and Pieter Abbeel.
\newblock A simple neural attentive meta-learner.
\newblock In \emph{ICLR}, 2018.

\bibitem[Ravi and Larochelle(2016)]{ravi2016optimization}
Sachin Ravi and Hugo Larochelle.
\newblock Optimization as a model for few-shot learning.
\newblock In \emph{International Conference on Learning Representations}, 2016.

\bibitem[Nichol et~al.(2018)Nichol, Achiam, and Schulman]{nichol2018first}
Alex Nichol, Joshua Achiam, and John Schulman.
\newblock On first-order meta-learning algorithms.
\newblock In \emph{arXiv:1803.02999}, 2018.

\bibitem[Liu et~al.(2019)Liu, Zhou, Long, Jiang, and Zhang]{liu2019learning}
Lu~Liu, Tianyi Zhou, Guodong Long, Jing Jiang, and Chengqi Zhang.
\newblock Learning to propagate for graph meta-learning.
\newblock In \emph{NeurIPS}, 2019.

\bibitem[Sung et~al.(2018)Sung, Yang, Zhang, Xiang, Torr, and
  Hospedales]{sung2018learning}
Flood Sung, Yongxin Yang, Li~Zhang, Tao Xiang, Philip~HS Torr, and Timothy~M
  Hospedales.
\newblock Learning to compare: relation network for few-shot learning.
\newblock In \emph{CVPR}, 2018.

\bibitem[Snell et~al.(2017)Snell, Swersky, and Zemel]{snell2017prototypical}
Jake Snell, Kevin Swersky, and Richard Zemel.
\newblock Prototypical networks for few-shot learning.
\newblock In \emph{NeurIPS}, 2017.

\bibitem[Finn et~al.(2017)Finn, Abbeel, and Levine]{finn2017model}
Chelsea Finn, Pieter Abbeel, and Sergey Levine.
\newblock Model-agnostic meta-learning for fast adaptation of deep networks.
\newblock In \emph{ICML}, 2017.

\bibitem[Wang et~al.(2022{\natexlab{b}})Wang, Dong, Huang, Chen, and
  Li]{wang2022faith}
Song Wang, Yushun Dong, Xiao Huang, Chen Chen, and Jundong Li.
\newblock Faith: Few-shot graph classification with hierarchical task graphs.
\newblock In \emph{IJCAI}, 2022{\natexlab{b}}.

\bibitem[Wang et~al.(2021)Wang, Huang, Chen, Wu, and Li]{wang2021reform}
Song Wang, Xiao Huang, Chen Chen, Liang Wu, and Jundong Li.
\newblock Reform: Error-aware few-shot knowledge graph completion.
\newblock In \emph{CIKM}, 2021.

\bibitem[Zhou et~al.(2019)Zhou, Cao, Zhang, Trajcevski, Zhong, and
  Geng]{zhou2019meta}
Fan Zhou, Chengtai Cao, Kunpeng Zhang, Goce Trajcevski, Ting Zhong, and
  Ji~Geng.
\newblock Meta-gnn: On few-shot node classification in graph meta-learning.
\newblock In \emph{CIKM}, 2019.

\bibitem[Liu et~al.(2021)Liu, Fang, Liu, and Hoi]{liu2021relative}
Zemin Liu, Yuan Fang, Chenghao Liu, and Steven~CH Hoi.
\newblock Relative and absolute location embedding for few-shot node
  classification on graph.
\newblock In \emph{AAAI}, 2021.

\bibitem[Tan et~al.(2022)Tan, Ding, Guo, and Liu]{tan2022graph}
Zhen Tan, Kaize Ding, Ruocheng Guo, and Huan Liu.
\newblock Graph few-shot class-incremental learning.
\newblock In \emph{WSDM}, 2022.

\bibitem[Liu et~al.(2022)Liu, Li, Li, Giunchiglia, Feng, and Guan]{liu2022few}
Yonghao Liu, Mengyu Li, Ximing Li, Fausto Giunchiglia, Xiaoyue Feng, and Renchu
  Guan.
\newblock Few-shot node classification on attributed networks with graph
  meta-learning.
\newblock In \emph{Proceedings of the 45th International ACM SIGIR Conference
  on Research and Development in Information Retrieval}, pages 471--481, 2022.

\bibitem[Wu et~al.()Wu, Zhou, Wen, Wan, Ma, Cheng, and Zhu]{wuinformation}
Zongqian Wu, Peng Zhou, Guoqiu Wen, Yingying Wan, Junbo Ma, Debo Cheng, and
  Xiaofeng Zhu.
\newblock Information augmentation for few-shot node classification.

\bibitem[Wang et~al.(2022{\natexlab{c}})Wang, Dong, Ding, Chen, and
  Li]{wang2022extreme}
Song Wang, Yushun Dong, Kaize Ding, Chen Chen, and Jundong Li.
\newblock Few-shot node classification with extremely weak supervision.
\newblock In \emph{WSDM}, 2022{\natexlab{c}}.

\bibitem[Xu et~al.(2021)Xu, Wang, Ni, Guo, and Tang]{xu2021self}
Minghao Xu, Hang Wang, Bingbing Ni, Hongyu Guo, and Jian Tang.
\newblock Self-supervised graph-level representation learning with local and
  global structure.
\newblock In \emph{International Conference on Machine Learning}, pages
  11548--11558. PMLR, 2021.

\bibitem[Suresh et~al.(2021)Suresh, Li, Hao, and
  Neville]{suresh2021adversarial}
Susheel Suresh, Pan Li, Cong Hao, and Jennifer Neville.
\newblock Adversarial graph augmentation to improve graph contrastive learning.
\newblock \emph{Advances in Neural Information Processing Systems},
  34:\penalty0 15920--15933, 2021.

\bibitem[Zhu et~al.(2021)Zhu, Xu, Liu, and Wu]{zhu2021empirical}
Yanqiao Zhu, Yichen Xu, Qiang Liu, and Shu Wu.
\newblock An empirical study of graph contrastive learning.
\newblock In \emph{Thirty-fifth Conference on Neural Information Processing
  Systems Datasets and Benchmarks Track (Round 2)}, 2021.

\bibitem[Khosla et~al.(2020)Khosla, Teterwak, Wang, Sarna, Tian, Isola,
  Maschinot, Liu, and Krishnan]{khosla2020supervised}
Prannay Khosla, Piotr Teterwak, Chen Wang, Aaron Sarna, Yonglong Tian, Phillip
  Isola, Aaron Maschinot, Ce~Liu, and Dilip Krishnan.
\newblock Supervised contrastive learning.
\newblock \emph{Advances in Neural Information Processing Systems}, 2020.

\bibitem[Akkas and Azad(2022)]{akkas2022jgcl}
Selahattin Akkas and Ariful Azad.
\newblock Jgcl: Joint self-supervised and supervised graph contrastive
  learning.
\newblock 2022.

\bibitem[Thakoor et~al.(2021{\natexlab{b}})Thakoor, Tallec, Azar, Munos,
  Veli{\v{c}}kovi{\'c}, and Valko]{thakoor2021bootstrapped}
Shantanu Thakoor, Corentin Tallec, Mohammad~Gheshlaghi Azar, Remi Munos, Petar
  Veli{\v{c}}kovi{\'c}, and Michal Valko.
\newblock Bootstrapped representation learning on graphs.
\newblock In \emph{ICLR Workshop on Geometrical and Topological Representation
  Learning}, 2021{\natexlab{b}}.

\bibitem[Bojchevski and G{\"u}nnemann(2018)]{bojchevski2018deep}
Aleksandar Bojchevski and Stephan G{\"u}nnemann.
\newblock Deep gaussian embedding of graphs: Unsupervised inductive learning
  via ranking.
\newblock In \emph{ICLR}, 2018.

\bibitem[Hu et~al.(2020)Hu, Fey, Zitnik, Dong, Ren, Liu, Catasta, and
  Leskovec]{hu2020open}
Weihua Hu, Matthias Fey, Marinka Zitnik, Yuxiao Dong, Hongyu Ren, Bowen Liu,
  Michele Catasta, and Jure Leskovec.
\newblock Open graph benchmark: Datasets for machine learning on graphs.
\newblock In \emph{NeurIPS}, 2020.

\bibitem[Shchur et~al.(2018)Shchur, Mumme, Bojchevski, and
  G{\"u}nnemann]{shchur2018pitfalls}
Oleksandr Shchur, Maximilian Mumme, Aleksandar Bojchevski, and Stephan
  G{\"u}nnemann.
\newblock Pitfalls of graph neural network evaluation.
\newblock \emph{Relational Representation Learning Workshop, NeurIPS 2018},
  2018.

\bibitem[Yang et~al.(2016)Yang, Cohen, and Salakhudinov]{yang2016revisiting}
Zhilin Yang, William Cohen, and Ruslan Salakhudinov.
\newblock Revisiting semi-supervised learning with graph embeddings.
\newblock In \emph{International conference on machine learning}, pages 40--48.
  PMLR, 2016.

\bibitem[Wang et~al.(2022{\natexlab{d}})Wang, Chen, and Li]{wang2022graph}
Song Wang, Chen Chen, and Jundong Li.
\newblock Graph few-shot learning with task-specific structures.
\newblock In \emph{NeurIPS}, 2022{\natexlab{d}}.

\bibitem[Grill et~al.(2020)Grill, Strub, Altch{\'e}, Tallec, Richemond,
  Buchatskaya, Doersch, Pires, Guo, Azar, et~al.]{grill2020bootstrap}
Jean-Bastien Grill, Florian Strub, Florent Altch{\'e}, Corentin Tallec, Pierre
  Richemond, Elena Buchatskaya, Carl Doersch, Bernardo Pires, Zhaohan Guo,
  Mohammad Azar, et~al.
\newblock Bootstrap your own latent: A new approach to self-supervised
  learning.
\newblock In \emph{NeurIPS}, 2020.

\bibitem[Fey and Lenssen(2019)]{Fey/Lenssen/2019}
Matthias Fey and Jan~E. Lenssen.
\newblock Fast graph representation learning with {PyTorch Geometric}.
\newblock In \emph{ICLR Workshop on Representation Learning on Graphs and
  Manifolds}, 2019.

\bibitem[Wang et~al.(2019)Wang, Zheng, Ye, Gan, Li, Song, Zhou, Ma, Yu, Gai,
  Xiao, He, Karypis, Li, and Zhang]{wang2019dgl}
Minjie Wang, Da~Zheng, Zihao Ye, Quan Gan, Mufei Li, Xiang Song, Jinjing Zhou,
  Chao Ma, Lingfan Yu, Yu~Gai, Tianjun Xiao, Tong He, George Karypis, Jinyang
  Li, and Zheng Zhang.
\newblock Deep graph library: A graph-centric, highly-performant package for
  graph neural networks.
\newblock \emph{arXiv preprint arXiv:1909.01315}, 2019.

\bibitem[Wang et~al.(2020{\natexlab{b}})Wang, Shen, Huang, Wu, Dong, and
  Kanakia]{wang2020microsoft}
Kuansan Wang, Zhihong Shen, Chiyuan Huang, Chieh-Han Wu, Yuxiao Dong, and
  Anshul Kanakia.
\newblock Microsoft academic graph: When experts are not enough.
\newblock \emph{Quantitative Science Studies}, 2020{\natexlab{b}}.

\bibitem[McAuley et~al.(2015)McAuley, Pandey, and
  Leskovec]{mcauley2015inferring}
Julian McAuley, Rahul Pandey, and Jure Leskovec.
\newblock Inferring networks of substitutable and complementary products.
\newblock In \emph{SIGKDD}, 2015.

\bibitem[Kingma and Ba(2015)]{kingma2014adam}
Diederik~P Kingma and Jimmy Ba.
\newblock Adam: A method for stochastic optimization.
\newblock In \emph{Proceedings of the 2015 International Conference on Learning
  Representations}, 2015.

\bibitem[Glorot and Bengio(2010)]{glorot2010understanding}
Xavier Glorot and Yoshua Bengio.
\newblock Understanding the difficulty of training deep feedforward neural
  networks.
\newblock In \emph{Proceedings of the thirteenth international conference on
  artificial intelligence and statistics}, 2010.

\bibitem[Paszke et~al.(2017)Paszke, Gross, Chintala, Chanan, Yang, DeVito, Lin,
  Desmaison, Antiga, and Lerer]{paszke2017automatic}
Adam Paszke, Sam Gross, Soumith Chintala, Gregory Chanan, Edward Yang, Zachary
  DeVito, Zeming Lin, Alban Desmaison, Luca Antiga, and Adam Lerer.
\newblock Automatic differentiation in pytorch.
\newblock In \emph{Proceedings of the 31st Conference on Neural Information
  Processing Systems}, 2017.

\end{thebibliography}
% For bibLaTeX users:
% \printbibliography
\newpage
\appendix
\section{Framework for Meta-learning Based FSNC Methods}

\begin{figure}[htbp]
  \centering\scalebox{0.8}{
  \includegraphics[width=\linewidth]{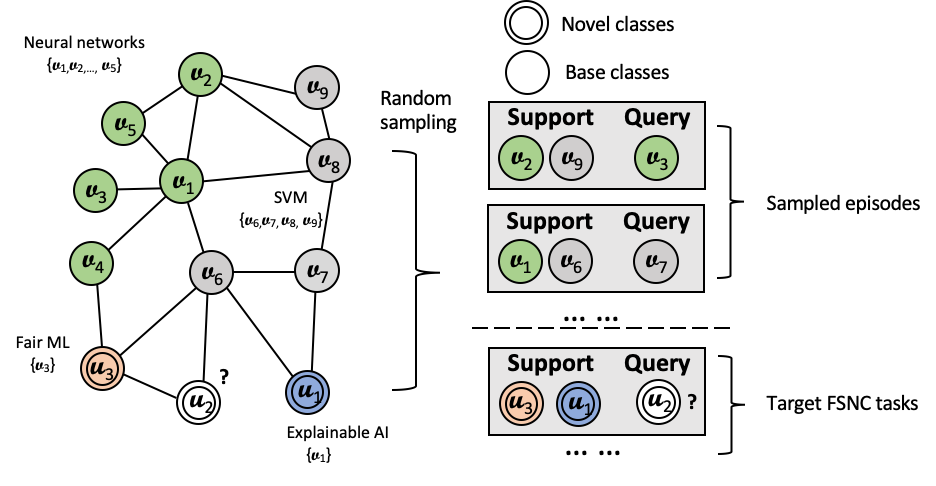}}
  \caption{The framework for meta-learning methods. Colors indicate different classes (e.g., \textit{Neural Networks, SVM, Fair ML, Explainable AI}). Specifically, white nodes denotes that the labels of those nodes are unavailable. Labels of all nodes in base classes are available. Different types of nodes indicate if nodes are from base classes or novel classes.}
  \label{fig:meta}
  \vspace{-0.0cm}
\end{figure}

\section{Framework for TLP with Self-Supervised GCL}
\label{app:TLP-self}

\begin{figure}[htbp]
  \centering\scalebox{0.95}{
  \includegraphics[width=\linewidth]{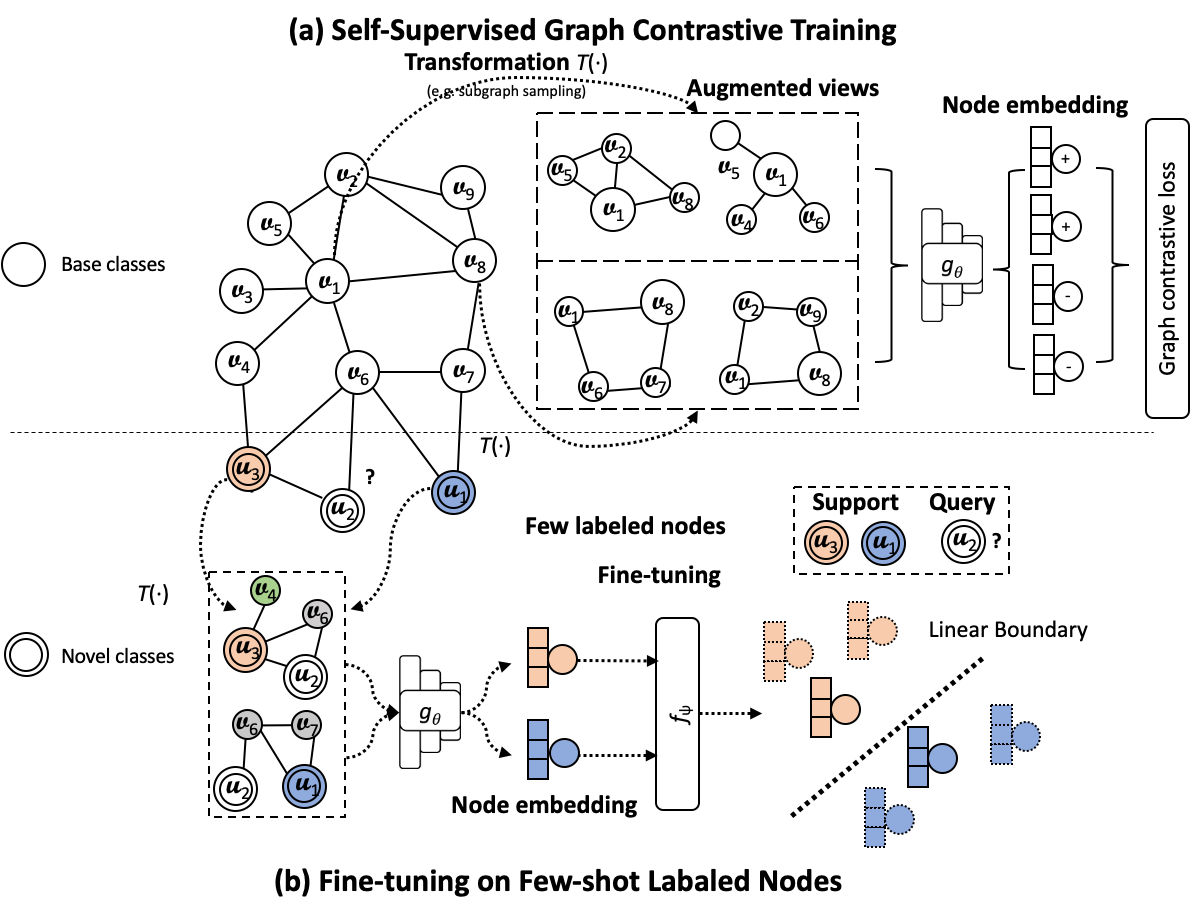}}
  \caption{The framework for TLP with self-supervised methods. Labels of all nodes in base classes are unavailable. Different types of nodes indicate if nodes are from base classes or novel classes.}
  \label{fig:meta}
  \vspace{-0.0cm}
\end{figure}

\newpage

\section{Default Values of Parameters in Evaluation Protocol}
In this section, we provide the default values of parameters used in our experiments. The details are provided in Table~\ref{tab:para}. It is noteworthy that the parameters are consistent for all models in both meta-learning and TLP methods. For the experiments that utilize a joint loss of TLP with self-supervised GCL and supervised GCL, we increase the patience number from $P$ to $2P$ to ensure convergence.
\label{app:val4protocol}
% Any possible appendices should be placed after bibliographies.
% If your paper has appendices, please submit the appendices together with the main body of the paper.
% There will be no separate supplementary material submission.
% The main text should be self-contained; reviewers are not obliged to look at the appendices when writing their review comments.

		\begin{table}[htbp]
	\setlength\tabcolsep{15pt}%调列距

		\centering
		\renewcommand{\arraystretch}{1.6}

		\caption{Default Values of Parameters in Evaluation Protocol for Experiments}
		\begin{tabular}{c|c|c}
\hline
\textbf{Parameters}&\textbf{Description}&\textbf{Value} \\\hline
$V$&  validation epoch interval & 10\\\hline
$I$& number of sampled meta-tasks for evaluation & 100\\\hline
$P$&patience number& 10 \\\hline
$E$& maximum epoch number& 10000 \\\hline
$R$& number of repeated experiments & 5 \\\hline
$N$& number of classes in each meta-task& 2,5 \\\hline
$K$& number of nodes for each class in each meta-task& 1,3,5 \\\hline
$M$& number of queries for each class in each meta-task& 10 \\ \hline

\end{tabular}
		\label{tab:para}
	\end{table}

% \newpage
\section{Description of Baselines}
In this section, we provide further details about the baselines used in our experiments.
\label{app_baseline}

\textit{Meta-learning} based methods:
\begin{itemize}
    \item \textbf{ProtoNet}~\cite{snell2017prototypical}: ProtoNet learns a prototype for each class in meta-tasks by averaging the embeddings of samples in this class. Then it conducts classification on query instances based on their distances to prototypes.
        \item \textbf{MAML}~\cite{finn2017model}: MAML first optimizes model parameters according to the gradients calculated on the support instances for several steps. Then it meta-updates parameters based on the loss of query instances calculated with the parameters updated on support instances.
        \item \textbf{Meta-GNN}~\cite{zhou2019meta}: Meta-GNN combines GNNs with the MAML strategy to apply meta-learning on graph-structured data. Specifically, Meta-GNN learns node embeddings with GNNs, while updating and meta-updating the GNN parameters based on the MAML strategy.
    \item \textbf{G-Meta}~\cite{huang2020graph}: G-Meta extracts a subgraph for each node to learn the node representation with GNNs. Then it conducts the classification on query nodes based on the MAML strategy to update and meta-update the parameters of GNNs.
    \item \textbf{GPN}~\cite{ding2020graph}: GPN proposes to learn node importance for each node in meta-tasks to select more beneficial nodes for classification. Then GPN utilizes ProtoNet to learn node prototypes via averaging node embeddings in a weighted manner.
    \item \textbf{AMM-GNN}~\cite{wang21AMM}: AMM-GNN proposes to extend MAML with an attribute matching mechanism. Specifically, the node embeddings will be adjusted according to the embeddings of nodes in the entire meta-task in an adaptive manner.
    \item \textbf{TENT}~\cite{wang2022task}: TENT reduces the variance among different meta-tasks for better generalization performance. In particular, TENT learns node and class representations by conducting node-level and class-level adaptations. It also incorporates task-level adaptations that maximizes the mutual information between the support set and the query set.
\end{itemize}
\textit{Transductive Linear Probing} with different  Pretraining methods:
\begin{itemize}
    \item \textbf{I-GNN}~\cite{tian2020rethinking}: I-GNN learns a GNN encoder with a classifier that is trained on all base classes $\mathbb{C}_{base}$ with the vanilla Cross-Entropy loss $L_{CE}$. Then for each meta-test task, the GNN will be frozen and a new classifier is learned based on the support set for classification.
    \item \textbf{MVGRL}~\cite{hassani2020contrastive}: MVGRL learns node
and graph level representations by contrasting the representations of two structural views of graphs, which include first-order
neighbors and a graph diffusion. It utilizes a Jensen-Shannon Divergence based contrastive loss $L_{JSD}$.
    \item \textbf{GraphCL}~\cite{you2020graph}: GraphCL proposes to leverage combinations of different transformations in GCL to facilitate GNNs with generalizability, transferrability, and robustness  without sophisticated architectures. It also uses $L_{JSD}$ as the objective.    \item \textbf{GRACE}~\cite{zhu2020deep}: GRACE proposes a hybrid scheme for generating different graph views on both structure and attribute levels. GRACE further provides theoretical
    justifications behind the motivation. It proposes a variant of Information Noise Contrastive Estimation $L_{InfoNCE}$ as the contrastive loss.
    \item \textbf{MERIT}~\cite{jin2021multi}: MERIT employs two different objectives named cross-view and cross-network contrastiveness to further maximize the agreement between node representations
across different views and networks. It uses $L_{InfoNCE}$ similar to that in GRACE as the loss function. 
    \item \textbf{SUGRL}~\cite{mo2022simple}: SUGRL proposes to simultaneously enlarge inter-class variation and reduce intra-class variation. The experimental results show promising improvements of generalization error with SUGRL. It also uses $L_{InfoNCE}$ similar to that in GRACE as the loss function.
    \item \textbf{BGRL}~\cite{thakoor2021bootstrapped}: BGRL leverages the concept of BYOL \cite{grill2020bootstrap} and applies it to graph-structured data by enforcing the agreement between positive views without any explicitly designs on negative views. Specially, it uses Mean Squared Error $L_{MSE}$ between positive views as the final loss.
    
\end{itemize}

% \newpage
\section{Statistics of Benchmark Datasets} \label{app:statistic}
\begin{table}[htbp]
	
	\setlength\tabcolsep{7.5pt}%调列距
\small
		\centering
		\renewcommand{\arraystretch}{1.6}

		\caption{Statistics of node classification datasets. }
		\begin{tabular}{cccccccc}
		\hline
        \textbf{Dataset}&\# Nodes & \# Edges & \# Features &$|\mathbb{C}|$& $|\mathbb{C}_{train}|$& $|\mathbb{C}_{dev}|$& $|\mathbb{C}_{test}|$\\
        \hline

    \texttt{CoraFull}&    19,793&63,421&8,710&70&40&15&15\\\hline
 \texttt{ogbn-arxiv}&169,343&1,166,243&128&40&20&10&10\\\hline
\texttt{Coauthor-CS}&18,333&81,894&6,805&15&5&5&5\\\hline
\texttt{Amazon-Computer}&13,752&245,861&767&10&4&3&3\\\hline
\texttt{Cora}&2,708&5,278&1,433&7&3&2&2\\\hline
 \texttt{CiteSeer}&3,327&4,552&3,703&6&2&2&2\\\hline

		\end{tabular}
		\label{tab:statistics}
	\end{table}

\section{Description of Benchmark Datasets}
\label{app:datasets}
In this section, we provide the detailed descriptions of the benchmark datasets used in our experiments. 
All the datasets are public and available on both PyTorch-Geometric~\cite{Fey/Lenssen/2019} and DGL~\cite{wang2019dgl}.
\begin{itemize}
        \item \textbf{CoraFull}~\cite{bojchevski2018deep} is a citation network that extends the prevalent small cora network. Specifically, it is achieved from the entire citation network, where nodes are papers, and edges denote the citation relations. The classes of nodes are obtained based on the paper topic. For this
dataset, we use 40/15/15 node classes for $\mathbb{C}_{train}$/$\mathbb{C}_{dev}$/$\mathbb{C}_{test}$.
    \item \textbf{ogbn-arxiv}~\cite{hu2020open} is a directed citation network that consists of CS papers from MAG~\cite{wang2020microsoft}. Here nodes represent CS arXiv papers, and edges denote the citation relations. The classes of nodes are assigned based on the 40 subject areas of CS papers in arXiv. For this
dataset, we use 20/10/10 node classes for $\mathbb{C}_{train}$/$\mathbb{C}_{dev}$/$\mathbb{C}_{test}$.
    \item \textbf{Coauthor-CS}~\cite{shchur2018pitfalls} is a  co-authorship graph based on the Microsoft Academic Graph from the KDD Cup 2016 challenge. Here, nodes are authors, and are connected by an edge if they co-authored a paper; node features represent paper keywords for each author’s papers, and class labels indicate most active fields of study for each author. For this
dataset, we use 5/5/5 node classes for $\mathbb{C}_{train}$/$\mathbb{C}_{dev}$/$\mathbb{C}_{test}$.
    \item \textbf{Amazon-Computer}~\cite{shchur2018pitfalls} includes segments of the Amazon co-purchase graph~\cite{mcauley2015inferring}, where nodes represent goods, edges indicate that two goods are frequently bought together, node features are bag-of-words encoded product reviews, and class labels are given by the product category. For this
dataset, we use 4/3/3 node classes for $\mathbb{C}_{train}$/$\mathbb{C}_{dev}$/$\mathbb{C}_{test}$.
    \item \textbf{Cora}~\cite{yang2016revisiting} is a citation network dataset where nodes mean paper and edges mean citation relationships. Each node has a predefined feature with 1,433 dimensions. The dataset is designed for the node classification task. The task is to predict the category of certain paper. For this
dataset, we use 3/2/2 node classes for $\mathbb{C}_{train}$/$\mathbb{C}_{dev}$/$\mathbb{C}_{test}$.
    \item \textbf{CiteSeer}~\cite{yang2016revisiting} is also a citation network dataset where nodes mean scientific publications and edges mean citation relationships. Each node has a predefined feature with 3,703 dimensions. The dataset is designed for the node classification task. The task is to predict the category of certain publication. For this
dataset, we use 2/2/2 node classes for $\mathbb{C}_{train}$/$\mathbb{C}_{dev}$/$\mathbb{C}_{test}$.
    
\end{itemize}

\section{Implementation Details}
In this section, we introduce the implementation details for all methods compared in our experiments. Specifically, for the encoders used in TLP methods, we follow the settings in the original papers of the corresponding models to ensure consistency, and we choose \textit{Logistic Regression} as the linear classifier for the final classification. For encoders in meta-learning methods, we utilize the original designs for papers using GNNs. For papers without using GNNs (i.e., ProtoNet~\cite{snell2017prototypical} and MAML~\cite{finn2017model}), we use a two-layer GCN~\cite{Kipf:2017tc} as the encoder with a hidden size of 16. We utilize the Adam optimizer~\cite{kingma2014adam} for all experiments with a learning rate of 0.001. To effectively initialize the GNNs in our experiments, we leverage the Xavier initialization~\cite{glorot2010understanding}. For meta-learning methods using the MAML framework, we set the number of meta-update steps as 20 with a meta-learning rate of 0.05. To ensure more stable convergence in meta-learning methods, we set the weight decay rate as $10^{-4}$. We set the dropout rate as 0.5 for better generalization performance. The evaluation protocol parameters are provided in Table~\ref{tab:para}. All experiments are implemented using PyTorch~\cite{paszke2017automatic}. We run all experiments on a single 80GB Nvidia A100 GPU.

% \newpage
\section{More Results}
\label{app:moreres}

\subsection{Visualization}
\begin{figure}[htbp]
		\centering
				\captionsetup[sub]{skip=-0.5pt}
	\subcaptionbox{GRACE$*$}{
\includegraphics[width=0.2\textwidth]{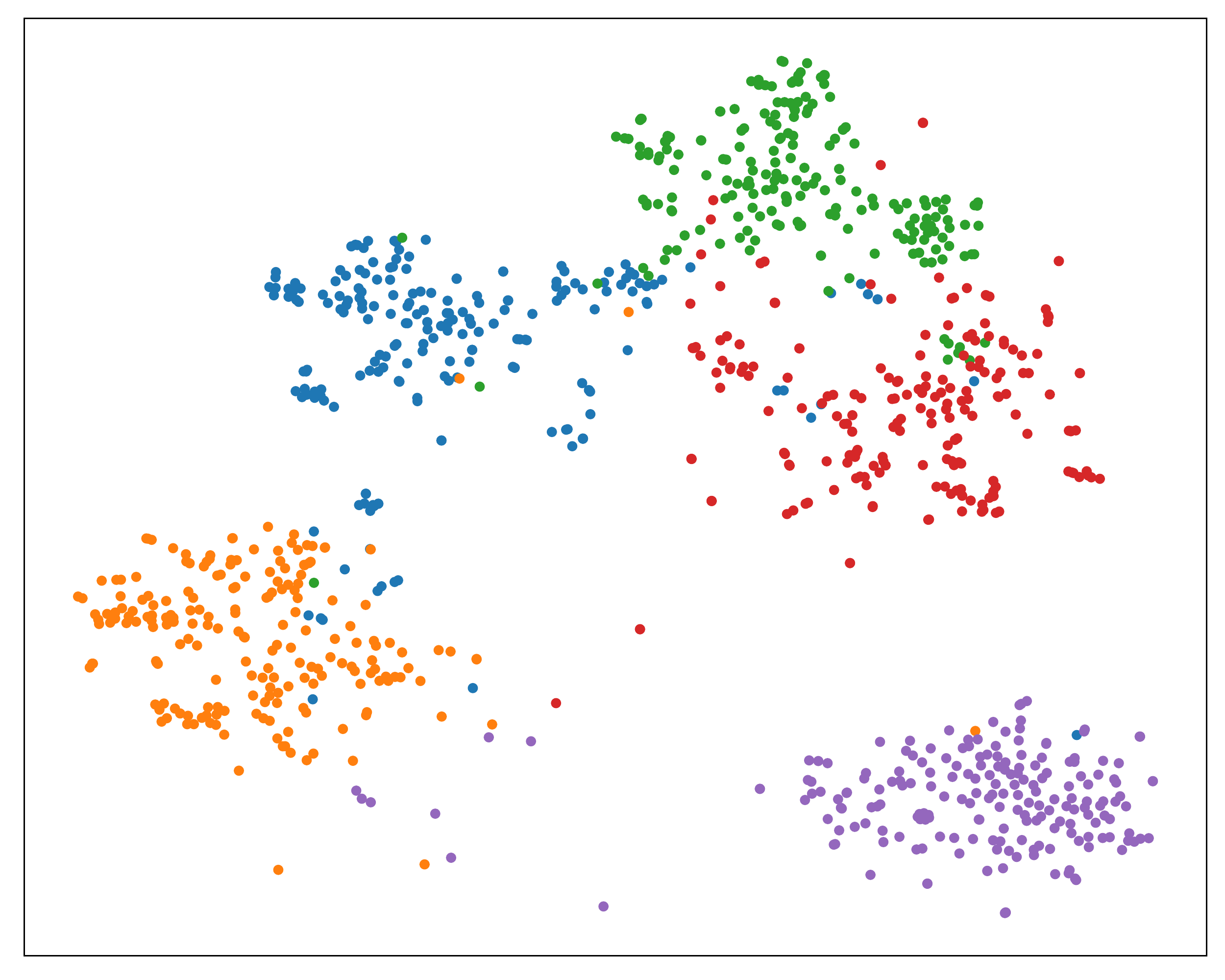}}
			\subcaptionbox{GRACE}{
\includegraphics[width=0.2\textwidth]{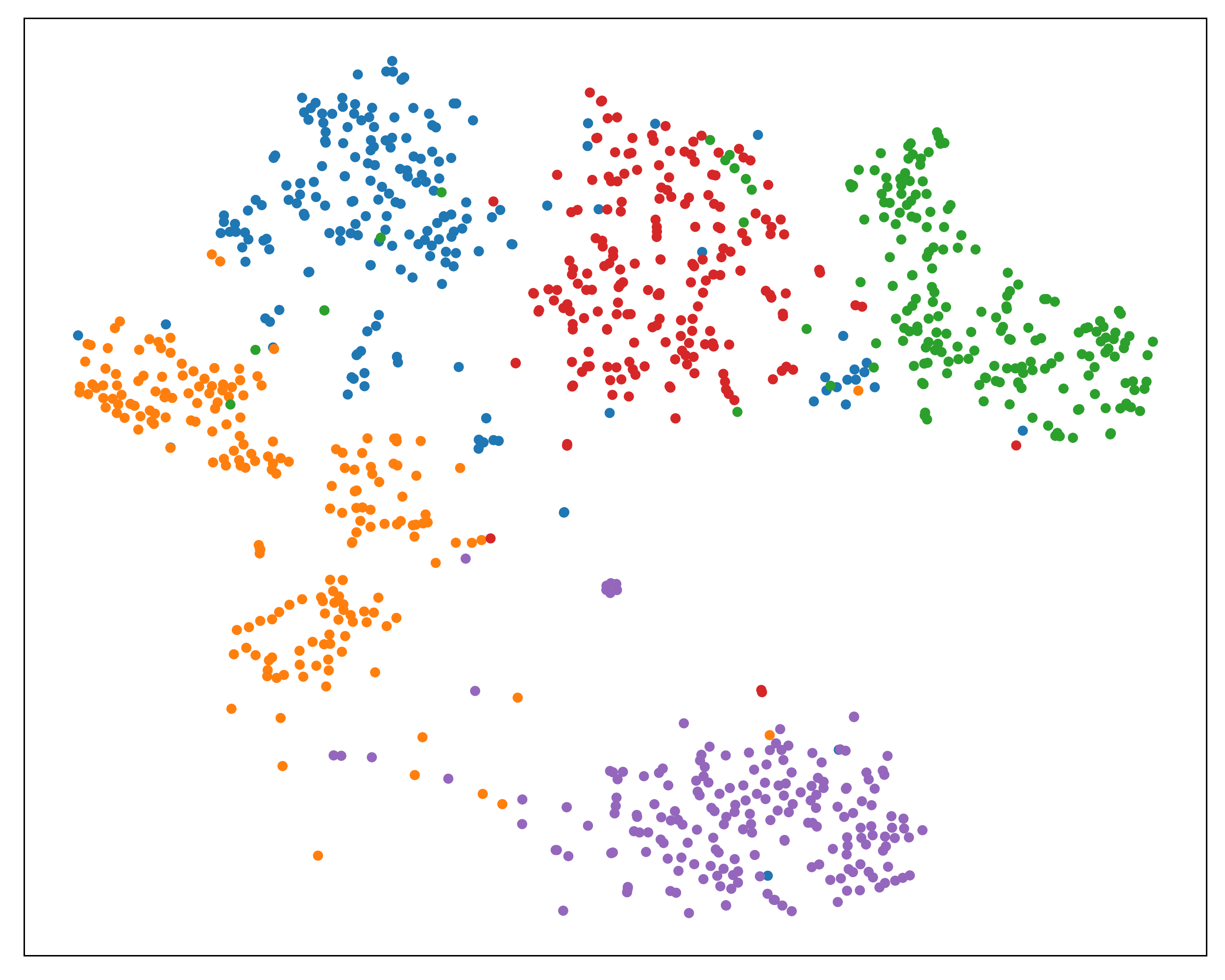}}
	\subcaptionbox{MERIT$*$}{
\includegraphics[width=0.2\textwidth]{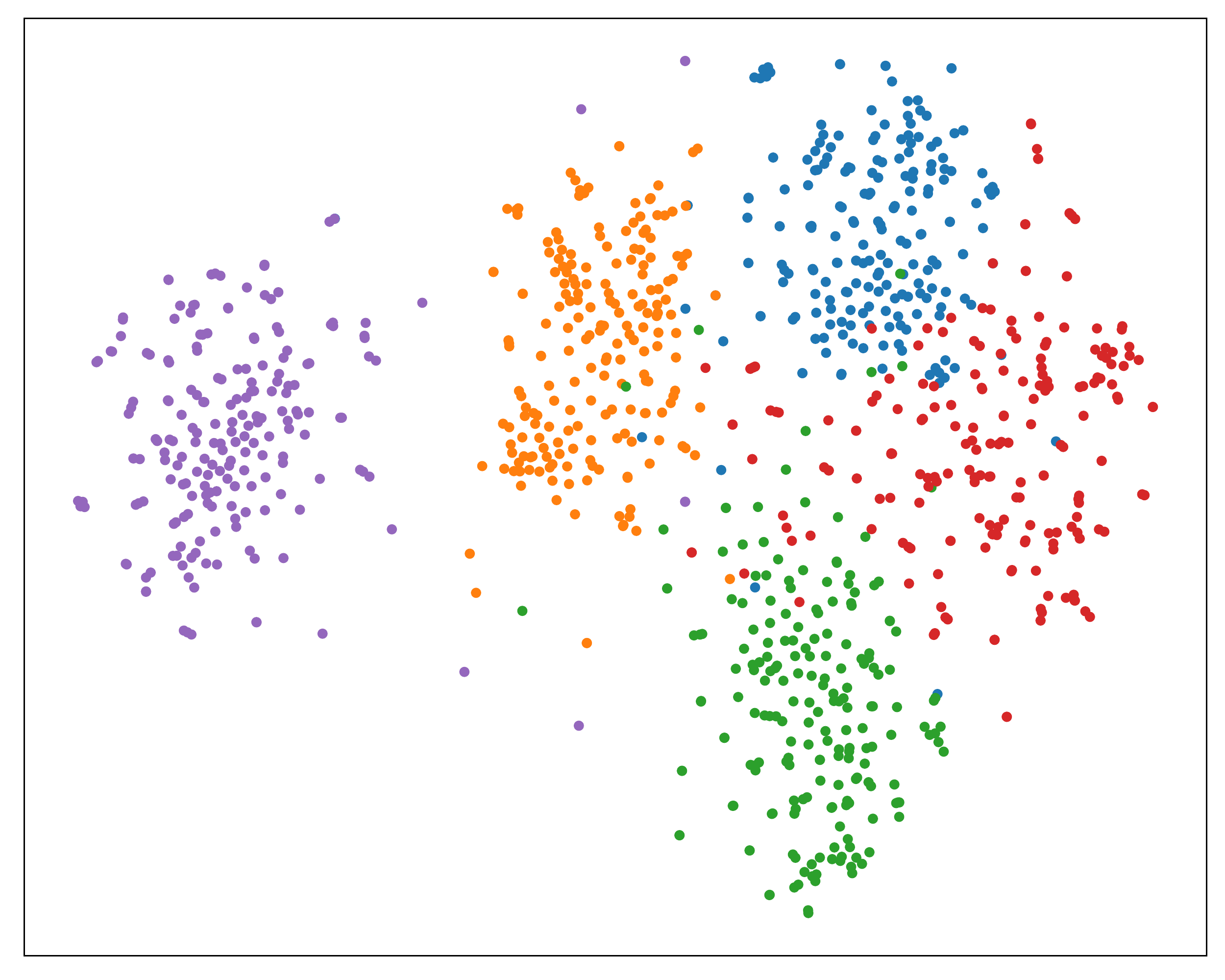}}
		\subcaptionbox{MERIT}{
\includegraphics[width=0.2\textwidth]{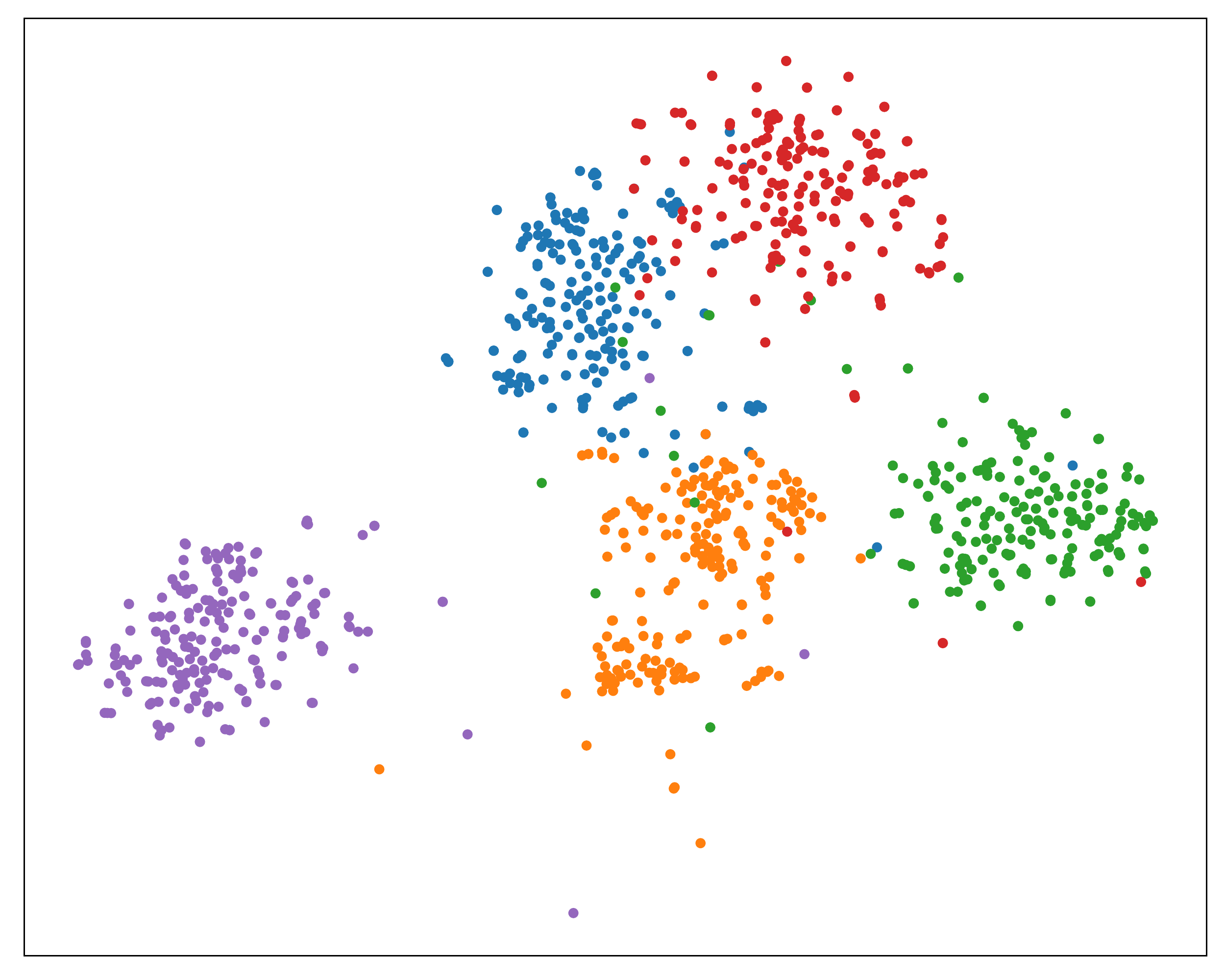}}

	\subcaptionbox{MVGRL$*$}{
\includegraphics[width=0.2\textwidth]{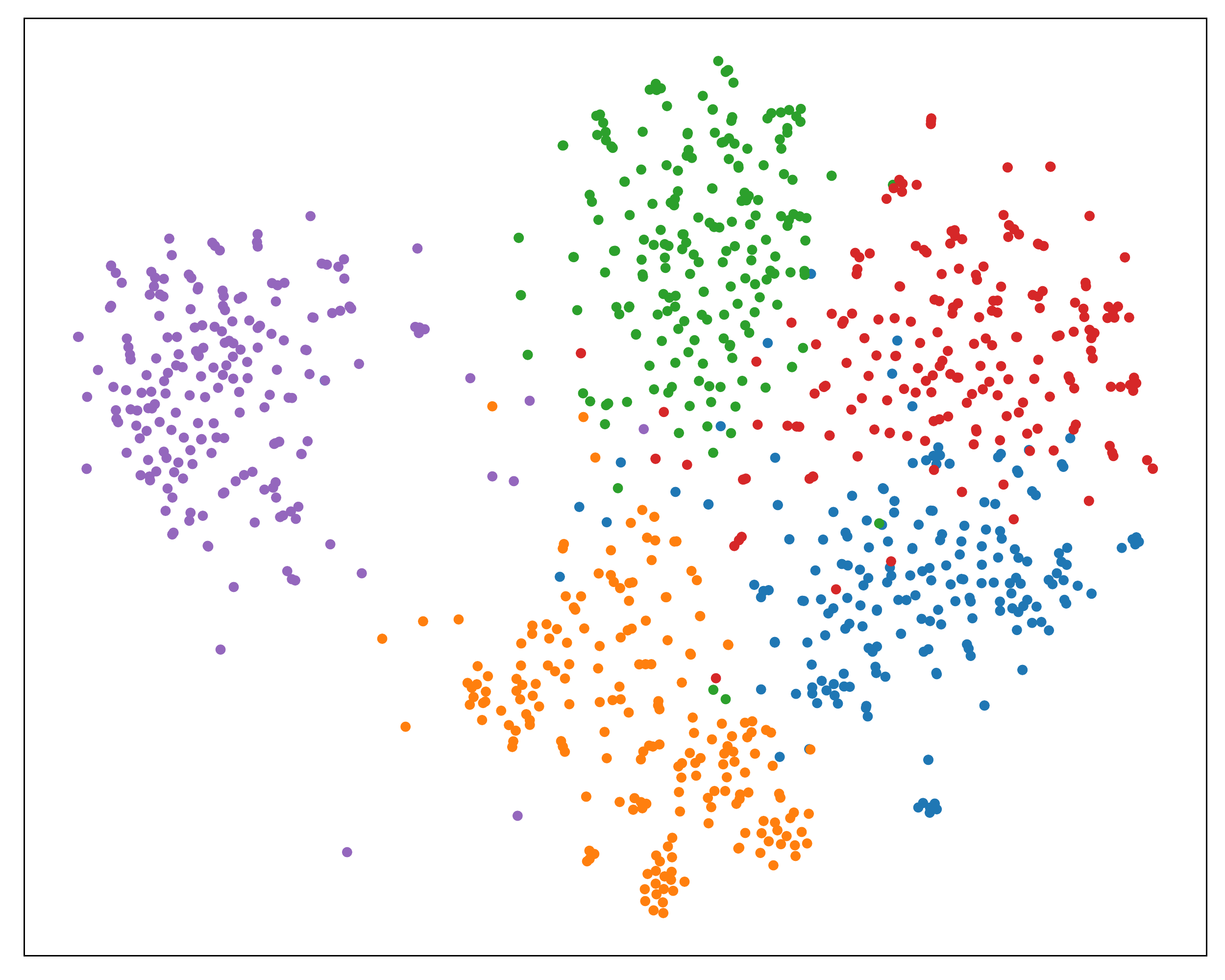}}
			\subcaptionbox{MVGRL}{
\includegraphics[width=0.2\textwidth]{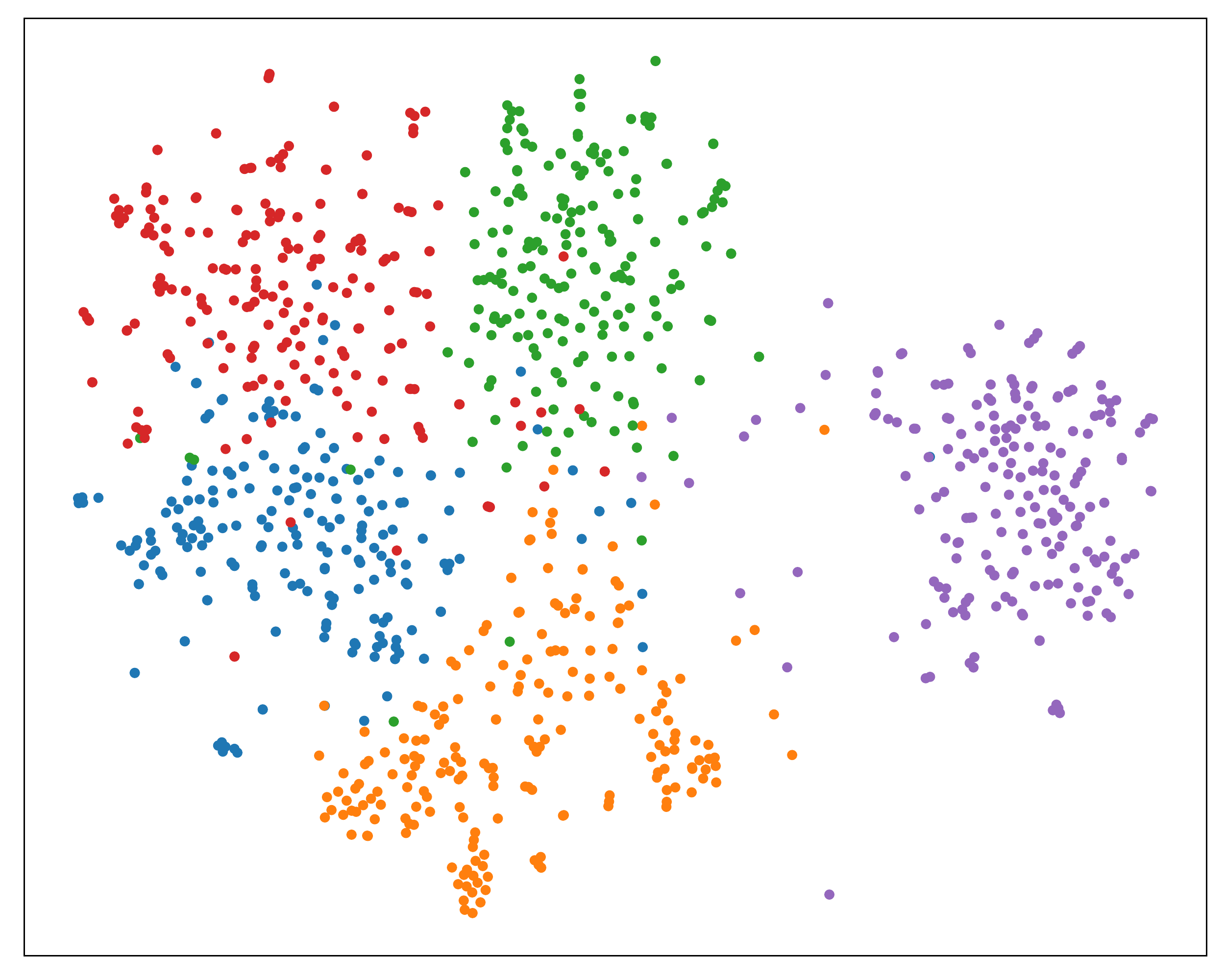}}
	\subcaptionbox{ProtoNet}{
\includegraphics[width=0.2\textwidth]{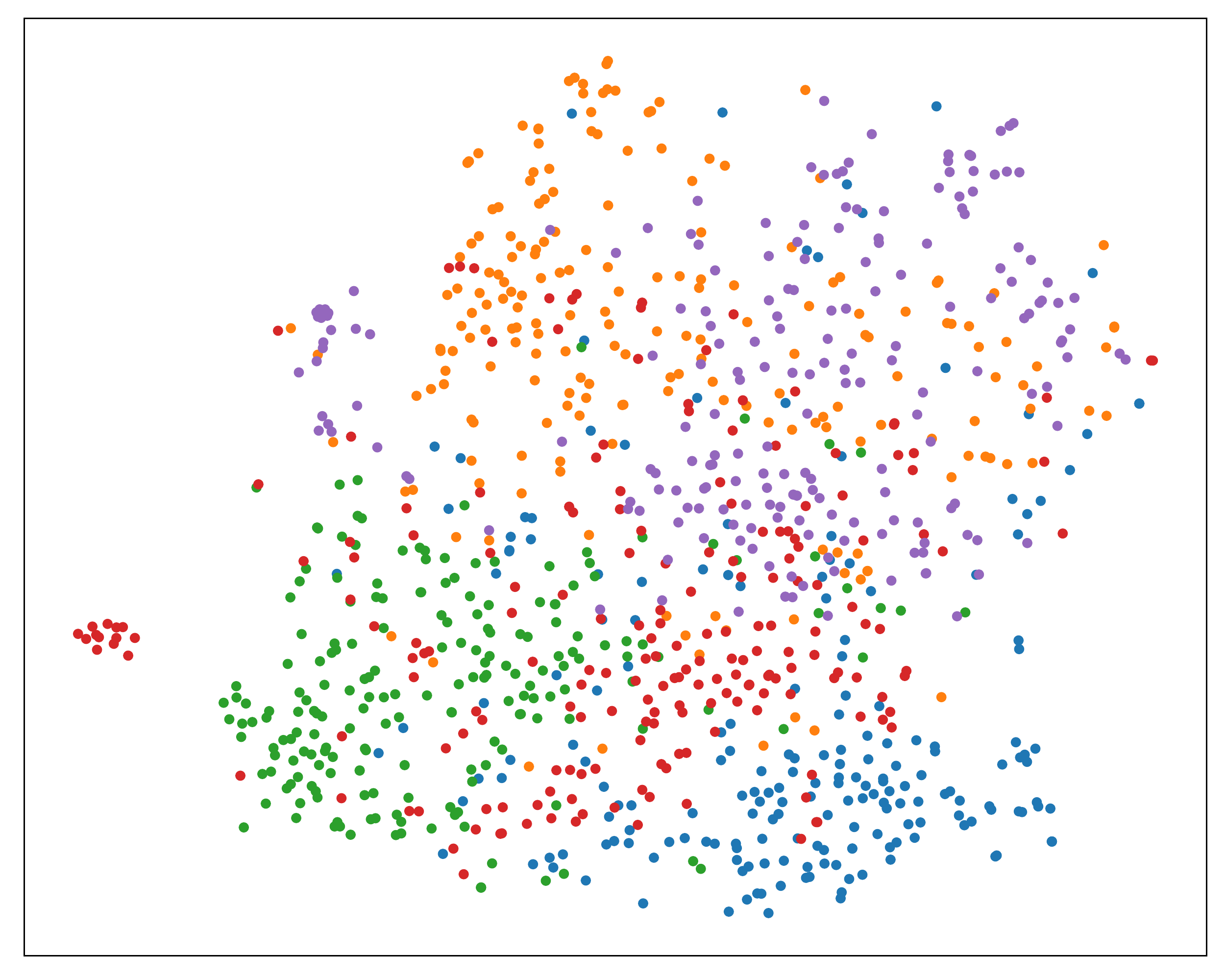}}
			\subcaptionbox{AMM-GNN}{
\includegraphics[width=0.2\textwidth]{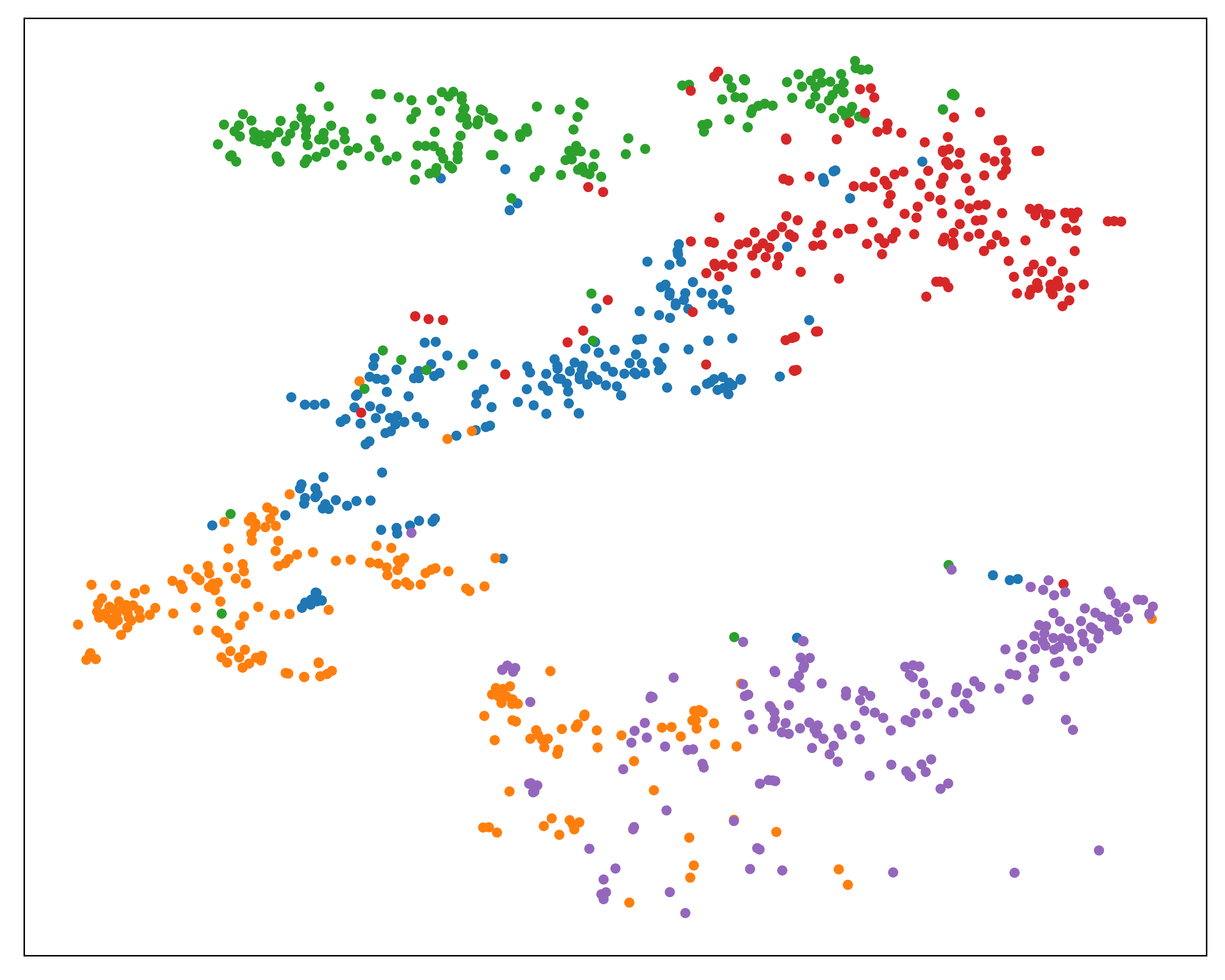}}

\caption{The t-SNE visualization results of meta-learning and TLP methods on \texttt{CoraFull}. TLP methods with $*$ are based on supervised GCL methods.}.
\label{fig:tsne_extra}
	\end{figure}

	\vspace{-0.in}
In this section, we provide additional visualization results for more meta-learning and TLP methods on \texttt{CoraFull} dataset in Fig.~\ref{fig:tsne_extra}.
% \newpage

\subsection{Node Representation Evaluation}
In this section, we provide the detailed node representation evaluations on two datasets \texttt{CoraFull} and \texttt{CiterSeer} based on NMI and ARI scores in Table~\ref{tab:nmi_result}. 
\label{app:evaluation}

	\begin{table*}[htbp]
		\setlength\tabcolsep{20pt}%调列距
	\small
		\centering
		\renewcommand{\arraystretch}{1.7}
		\caption{The overall NMI ($\uparrow$) and ARI ($\uparrow$) results of meta-learning and TLP methods on two datasets}
        \vspace{-0.0in}
        \scalebox{0.83}{
		\begin{tabular}{c||c|c||c|c}
			\hline
			%\multirow{2}{*}{Model}
			Dataset&\multicolumn{2}{c||}{\texttt{CoraFull}}&\multicolumn{2}{c}{\texttt{CiteSeer}}
			\\
			\hline
						Metrics&\multicolumn{1}{c|}{NMI}&\multicolumn{1}{c||}{ARI}&\multicolumn{1}{c|}{NMI}&\multicolumn{1}{c}{ARI}\\

					\hline
			\multicolumn{5}{c}{Meta-learning} \\\hline	
			MAML&$0.1622$&$0.0597$&$0.0754$&$0.0602$\\\hline
			ProtoNet&$0.2669$&$0.1263$&$0.0915$&$0.0765$\\\hline
			AMM-GNN&$0.6247$&$0.5087$&$0.2090$&$0.1781$\\\hline
			G-Meta&$0.5003$&$0.3702$&$0.1913$&$0.1502$\\\hline
			Meta-GNN&$0.5534$&$0.4196$&$0.1317$&$0.1171$\\\hline
			GPN&$0.6001$&$0.4599$&$0.2119$&$0.2087$\\\hline
			TENT&$0.5760$&$0.4652$&$0.0930$&$0.0811$\\\hline

			\multicolumn{5}{c}{Supervised GCL} \\\hline
GRACE&$0.7199$&$0.6239$&$0.4693$&$0.4769$\\\hline
MERIT&$0.6119$&$0.4470$&$0.3471$&$0.3482$\\\hline
GraphCL&$0.2474$&$0.0852$&$0.1321$&$0.0711$\\\hline
SUGRL&$0.7298$&$0.6626$&$0.3927$&$0.4451$\\\hline
MVGRL&$0.6412$&$0.5038$&$0.2445$&$0.2146$\\\hline

			\multicolumn{5}{c}{Self-supervised GCL} \\\hline
			GRACE&$0.6781$&$0.5856$&$0.2663$&$0.2778$\\\hline
MERIT&$0.7419$&$0.6590$&$0.3923$&$0.4014$\\\hline
GraphCL&$0.7023$&$0.5628$&$0.5579$&$0.5890$\\\hline
SUGRL&$0.7680$&$0.7049$&$0.3952$&$0.4460$\\\hline
MVGRL&$0.6227$&$0.4788$&$0.2554$&$0.2232$\\\hline
	\hline
				
\end{tabular}}
		\label{tab:nmi_result}
	\end{table*}
	
\newpage
\subsection{Main Results for the Other Three Datasets or Other Settings}
In this section, we further provide results for the other three datasets used in our experiments: \texttt{Coauthor-CS}, \texttt{Amazon-Computer}, and \texttt{Cora}, and 2-way classification results on \texttt{CoraFull}, \texttt{ogbn-arxiv}, and \texttt{Coauthor-CS}: 
	\begin{table*}[htbp]
		\setlength\tabcolsep{4.5pt}%调列距
	\scriptsize
		\centering
		\renewcommand{\arraystretch}{1.8}

		\caption{The overall few-shot node classification results of meta-learning methods and TLP with different GCL methods under different settings. Accuracy ($\uparrow$) and confidence interval ($\downarrow$) are in $\%$. The best and second best results are \textbf{bold} and \underline{underlined}, respectively.}\label{tab:otherdata}
		\vspace{0.in}
		\begin{tabular}{c||c|c||c|c||c|c}
			\hline
			%\multirow{2}{*}{Model}
			Dataset&\multicolumn{2}{c||}{\texttt{Coauthor-CS}}&\multicolumn{2}{c||}{\texttt{Amazon-Computer}}&\multicolumn{2}{c}{\texttt{Cora}}
			\\
			\hline
						Setting&\multicolumn{1}{c|}{5-way 1-shot}&\multicolumn{1}{c||}{5-way 5-shot}&\multicolumn{1}{c|}{2-way 1-shot}&\multicolumn{1}{c||}{2-way 5-shot}&\multicolumn{1}{c|}{2-way 1-shot}&\multicolumn{1}{c}{2-way 5-shot}\\
						
\hline \multicolumn{7}{c}{Meta-learning}\\\hline
MAML&$27.98\pm1.42$&$42.12\pm1.40$&$52.67\pm2.11$&$58.23\pm2.53$&$53.13\pm2.26$&$57.39\pm2.23$\\\hline
ProtoNet&$32.13\pm1.52$&$49.25\pm1.50$&$61.98\pm2.95$&$70.20\pm2.64$&$53.04\pm2.36$&$57.92\pm2.34$\\\hline
Meta-GNN&$52.86\pm2.14$&$68.59\pm1.49$&$65.19\pm3.29$&$78.65\pm3.12$&$\underline{65.27\pm2.93}$&$72.51\pm1.91$\\\hline
GPN&$60.66\pm2.07$&$\mathbf{81.79\pm1.18}$&$57.26\pm1.50$&$77.63\pm2.91$&$62.61\pm2.71$&$76.39\pm2.33$\\\hline
AMM-GNN&$\underline{62.04\pm2.26}$&$\underline{81.78\pm1.24}$&$\underline{71.04\pm3.56}$&$\underline{79.21\pm3.38}$&$65.23\pm2.67$&$\mathbf{82.30\pm2.07}$\\\hline
G-Meta&$59.68\pm2.16$&$74.18\pm1.29$&$63.68\pm3.05$&$70.21\pm3.16$&$\mathbf{67.03\pm3.22}$&$\underline{80.05\pm1.98}$\\\hline
TENT&$\mathbf{63.70\pm1.88}$&$76.90\pm1.19$&$\mathbf{71.15\pm3.11}$&$\mathbf{79.25\pm2.61}$&$53.05\pm2.78$&$62.15\pm2.13$\\\hline
	\hline \multicolumn{7}{c}{TLP with Supervised GCL}\\\hline
I-GNN&$43.89\pm1.82$&$55.93\pm1.46$&$62.32\pm2.89$&$72.81\pm2.93$&$54.45\pm3.13$&$65.18\pm2.21$\\\hline
MVGRL&$62.16\pm2.05$&$84.79\pm1.13$&$64.69\pm2.84$&$84.84\pm2.10$&$57.24\pm2.07$&$78.04\pm2.08$\\\hline
GraphCL&$54.72\pm2.62$&$84.02\pm1.23$&$\mathbf{75.65\pm3.05}$&$\underline{88.31\pm1.86}$&$57.10\pm2.27$&$79.53\pm1.98$\\\hline
GRACE&$\underline{76.48\pm1.95}$&$90.22\pm0.84$&$\underline{75.57\pm3.01}$&$87.69\pm2.17$&$\mathbf{66.79\pm2.96}$&$\underline{89.77\pm1.59}$\\\hline
MERIT&$71.70\pm2.88$&$\underline{91.54\pm0.75}$&$72.10\pm3.86$&$\mathbf{94.56\pm1.19}$&$\underline{65.29\pm3.23}$&$\mathbf{91.02\pm2.00}$\\\hline
SUGRL&$\mathbf{84.78\pm1.47}$&$\mathbf{93.01\pm0.62}$&$71.42\pm2.68$&$84.12\pm0.75$&$53.21\pm1.80$&$57.64\pm1.79$\\\hline
	\hline \multicolumn{7}{c}{TLP with Self-supervised GCL}\\\hline
MVGRL&$67.51\pm2.21$&$88.72\pm1.04$&$66.49\pm2.75$&$86.31\pm2.09$&$71.17\pm3.04$&$89.91\pm1.45$\\\hline
GraphCL&$70.26\pm2.19$&$87.32\pm1.19$&$77.26\pm3.12$&$94.13\pm1.34$&$\underline{73.51\pm3.18}$&$\underline{92.38\pm1.30}$\\\hline
BGRL&$64.72\pm2.35$&$90.10\pm0.88$&$68.58\pm3.06$&$89.15\pm1.97$&$60.14\pm2.33$&$79.86\pm1.92$\\\hline
GRACE&$79.38\pm1.75$&$91.68\pm0.72$&$75.23\pm2.59$&$90.48\pm1.24$&$71.21\pm2.97$&$89.68\pm1.65$\\\hline
MERIT&$\underline{85.74\pm1.70}$&$\underline{95.78\pm0.61}$&$\underline{78.14\pm3.82}$&$\underline{95.98\pm1.38}$&$67.67\pm2.99$&$\mathbf{95.42\pm1.21}$\\\hline
SUGRL&$\mathbf{91.63\pm1.22}$&$\mathbf{96.30\pm0.51}$&$\mathbf{85.05\pm2.23}$&$\mathbf{97.15\pm0.81}$&$\mathbf{82.35\pm2.21}$&$92.22\pm1.15$\\\hline

\end{tabular}
		\label{tab:app_result}
	\end{table*}
	
\begin{table*}[htbp]
		\setlength\tabcolsep{4.5pt}%调列距
	\scriptsize
		\centering
		\renewcommand{\arraystretch}{1.8}

		\caption{The overall few-shot node classification results of meta-learning methods and TLP with different GCL methods under different settings. Accuracy ($\uparrow$) and confidence interval ($\downarrow$) are in $\%$. The best and second best results are \textbf{bold} and \underline{underlined}, respectively.}\label{tab:otherdata}
		\vspace{0.in}
		\begin{tabular}{c||c|c||c|c||c|c}
			\hline
			%\multirow{2}{*}{Model}
			Dataset&\multicolumn{2}{c||}{\texttt{CoraFull}}&\multicolumn{2}{c||}{\texttt{ogbn-arxiv}}&\multicolumn{2}{c}{\texttt{Coauthor-CS}}
			\\
			\hline
						Setting&\multicolumn{1}{c|}{2-way 1-shot}&\multicolumn{1}{c||}{2-way 5-shot}&\multicolumn{1}{c|}{2-way 1-shot}&\multicolumn{1}{c||}{2-way 5-shot}&\multicolumn{1}{c|}{2-way 1-shot}&\multicolumn{1}{c}{2-way 5-shot}\\
	\hline \multicolumn{7}{c}{Meta-learning}\\\hline
	
MAML&$50.90\pm2.30$&$56.19\pm2.37$&$58.16\pm2.35$&$65.10\pm2.56$&$56.90\pm2.41$&$66.78\pm2.35$\\\hline
ProtoNet&$57.10\pm2.47$&$72.71\pm2.55$&$62.56\pm2.86$&$75.82\pm2.79$&$59.92\pm2.70$&$71.69\pm2.51$\\\hline
Meta-GNN&$75.28\pm3.85$&$84.59\pm2.89$&$62.52\pm3.41$&$70.15\pm2.68$&$85.90\pm2.96$&$90.11\pm2.17$\\\hline
GPN&$74.29\pm3.47$&$85.58\pm2.53$&$64.00\pm3.71$&$76.78\pm3.50$&$84.31\pm2.73$&$90.36\pm1.90$\\\hline
AMM-GNN&$77.29\pm3.40$&$\underline{88.66\pm2.06}$&$\underline{64.68\pm3.13}$&$\underline{78.42\pm2.71}$&$\underline{84.38\pm2.85}$&$\mathbf{94.74\pm1.20}$\\\hline
G-Meta&$\mathbf{78.23\pm3.41}$&$\mathbf{89.49\pm2.04}$&$63.03\pm3.32$&$76.56\pm2.89$&$84.19\pm2.97$&$91.02\pm1.61$\\\hline
TENT&$\underline{77.75\pm3.29}$&$88.20\pm2.61$&$\mathbf{70.30\pm2.85}$&$\mathbf{81.35\pm2.77}$&$\mathbf{87.85\pm2.48}$&$\underline{91.75\pm1.60}$\\\hline
	\hline \multicolumn{7}{c}{Supervised GCL}\\\hline
I-GNN&$68.43\pm2.94$&$78.20\pm2.83$&$\underline{65.21\pm2.86}$&$\underline{77.10\pm2.46}$&$65.35\pm3.09$&$76.83\pm2.48$\\\hline
MVGRL&$65.62\pm3.11$&$84.41\pm2.35$&OOM&OOM&$78.08\pm3.59$&$91.78\pm1.66$\\\hline
GraphCL&$60.81\pm2.23$&$81.25\pm2.29$&OOM&OOM&$74.16\pm2.88$&$88.43\pm1.73$\\\hline
GRACE&$\mathbf{76.78\pm3.49}$&$\mathbf{93.62\pm1.32}$&OOM&OOM&$\underline{86.22\pm2.53}$&$94.11\pm1.27$\\\hline
MERIT&$75.52\pm6.53$&$88.03\pm5.11$&OOM&OOM&$77.52\pm7.58$&$\mathbf{96.62\pm2.12}$\\\hline
SUGRL&$\underline{75.98\pm2.98}$&$\underline{90.02\pm1.53}$&$\mathbf{73.48\pm2.55}$&$\mathbf{81.04\pm1.68}$&$\mathbf{88.45\pm1.62}$&$\underline{95.10\pm0.56}$\\\hline
	\hline \multicolumn{7}{c}{Self-supervised GCL}\\\hline
MVGRL&$78.81\pm3.32$&$91.03\pm1.80$&OOM&OOM&$78.59\pm2.92$&$93.54\pm1.40$\\\hline
GraphCL&$78.49\pm3.26$&$91.32\pm2.11$&OOM&OOM&$78.51\pm3.12$&$91.34\pm1.57$\\\hline
BGRL&$61.08\pm2.65$&$85.03\pm2.25$&$\underline{59.91\pm2.36}$&$\underline{76.75\pm0.86}$&$76.85\pm3.23$&$94.69\pm1.29$\\\hline
GRACE&$\underline{82.80\pm3.13}$&$93.06\pm2.17$&OOM&OOM&$89.46\pm2.26$&$95.53\pm1.05$\\\hline
MERIT&$77.46\pm3.14$&$\underline{94.65\pm1.31}$&OOM&OOM&$\underline{94.31\pm1.73}$&$\underline{98.35\pm0.57}$\\\hline
SUGRL&$\mathbf{87.98\pm2.72}$&$\mathbf{95.81\pm1.69}$&$\mathbf{82.45\pm2.94}$&$\mathbf{91.68\pm1.57}$&$\mathbf{96.81\pm1.31}$&$\mathbf{98.90\pm0.48}$\\\hline

\end{tabular}
		\label{tab:app_result}
	\end{table*}

\newpage

		\begin{figure}[htbp]
		\centering
\includegraphics[width=0.98\textwidth]{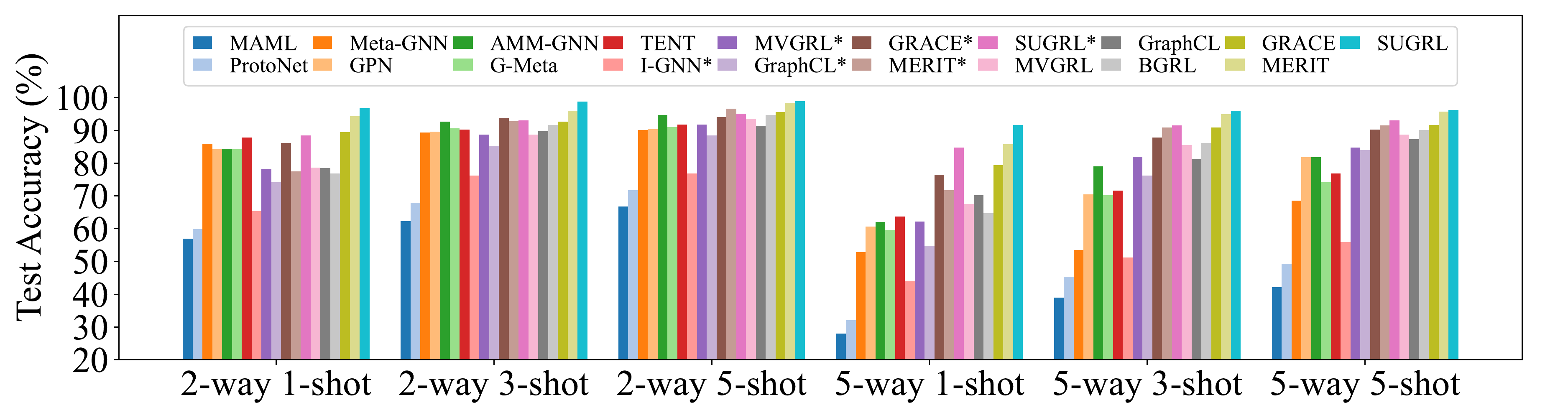}
	%\subcaptionbox{\includegraphics[width=0.48\textwidth]{img/nwaykshot.pdf}}

\caption{$N$-way $K$-shot results on \texttt{Coauthor-CS}, meta-learning and TLP. TLP Methods with $\ast$ are based on supervised GCL methods and I-GNN.}
	\end{figure}	
	
	\newpage
	
		\begin{figure}[htbp]
		\centering

\includegraphics[width=0.98\textwidth]{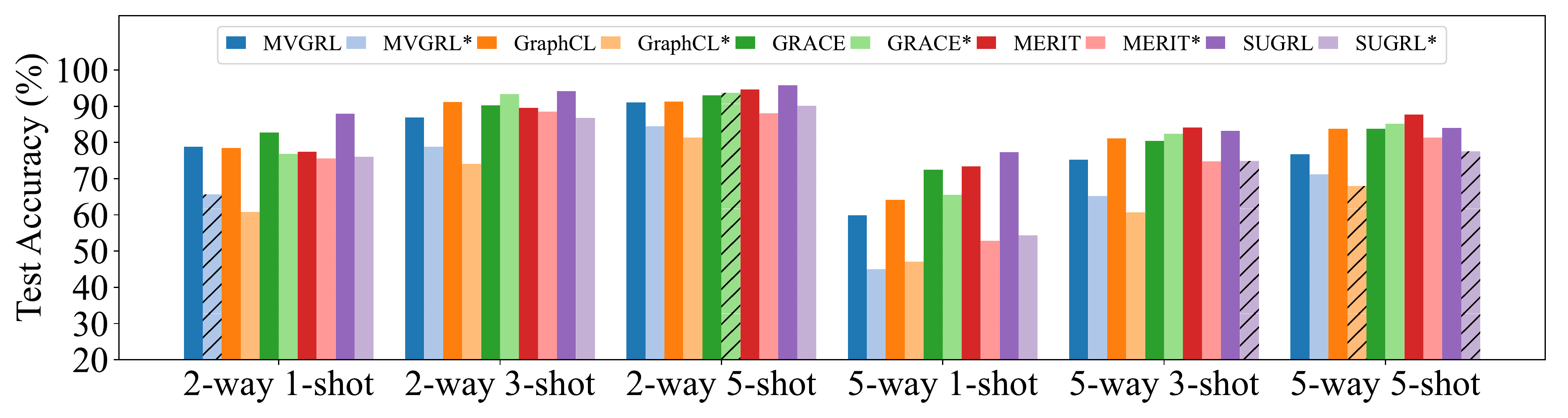}
	%\subcaptionbox{\includegraphics[width=0.48\textwidth]{img/nwaykshot.pdf}}

\caption{$N$-way $K$-shot results on \texttt{CoraFull}, TLP with self-supervised and supervised GCL. TLP Methods with $\ast$ are based on supervised GCL methods.}
	\end{figure}
		\begin{figure}[htbp]
		\centering
\includegraphics[width=0.98\textwidth]{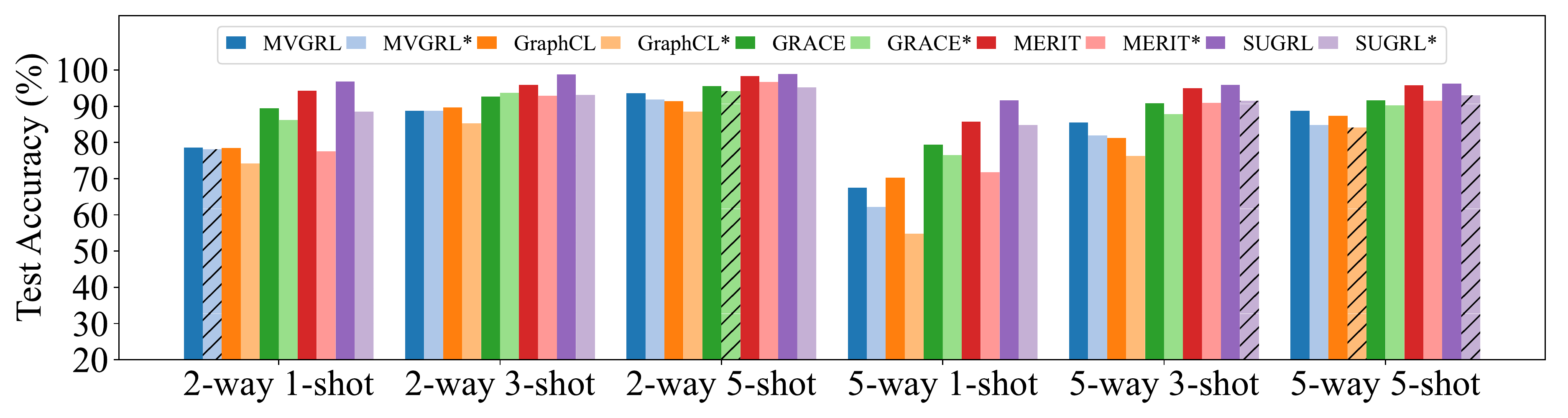}
	%\subcaptionbox{\includegraphics[width=0.48\textwidth]{img/nwaykshot.pdf}}

\caption{$N$-way $K$-shot results on \texttt{Coauthor-CS}, TLP with self-supervised and supervised GCL. TLP Methods with $\ast$ are based on supervised GCL methods.}
	\end{figure}	

			\begin{figure}[htbp]
		\centering
		\captionsetup[sub]{skip=-1pt}
		\subcaptionbox{Results on dataset \texttt{Amazon-Computer}}
		{\includegraphics[width=0.48\textwidth]{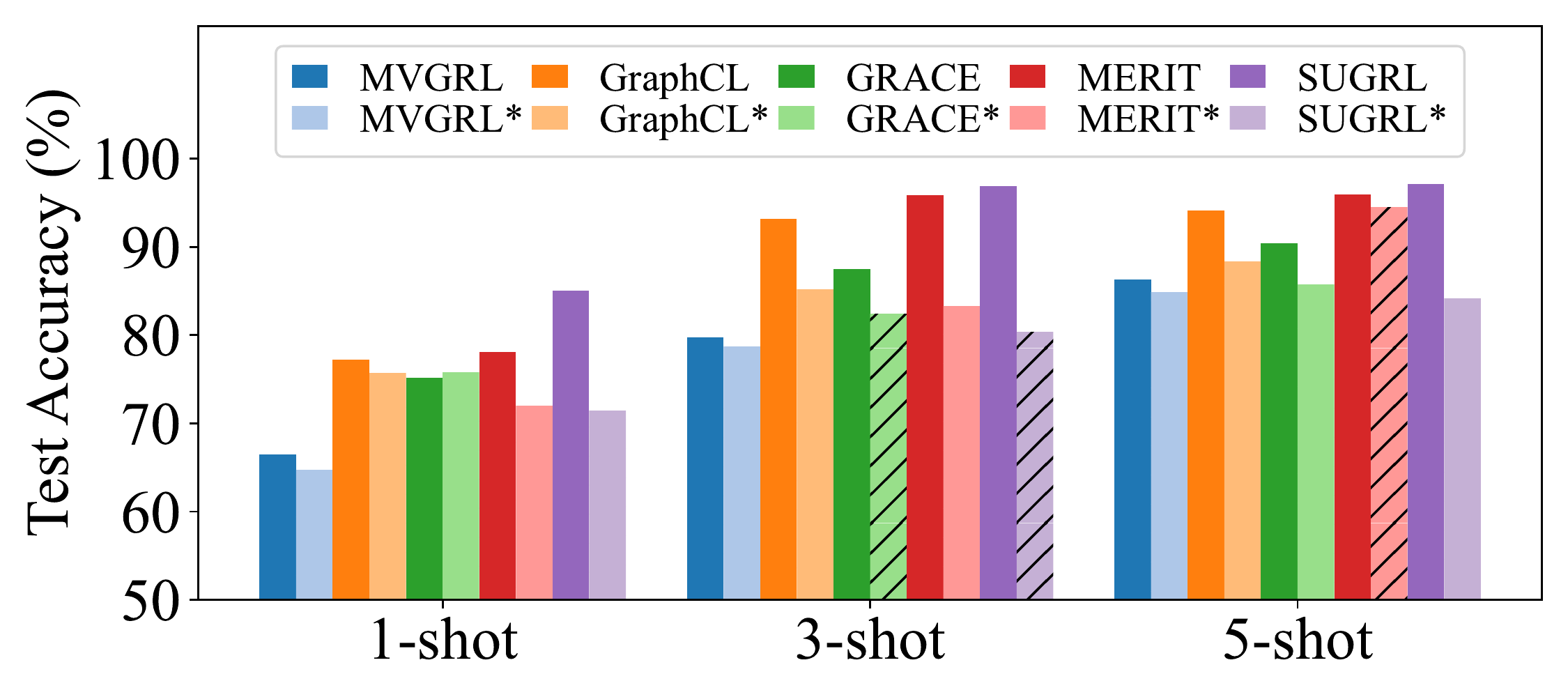}}
		\subcaptionbox{Results on dataset \texttt{CiteSeer}}
{\includegraphics[width=0.48\textwidth]{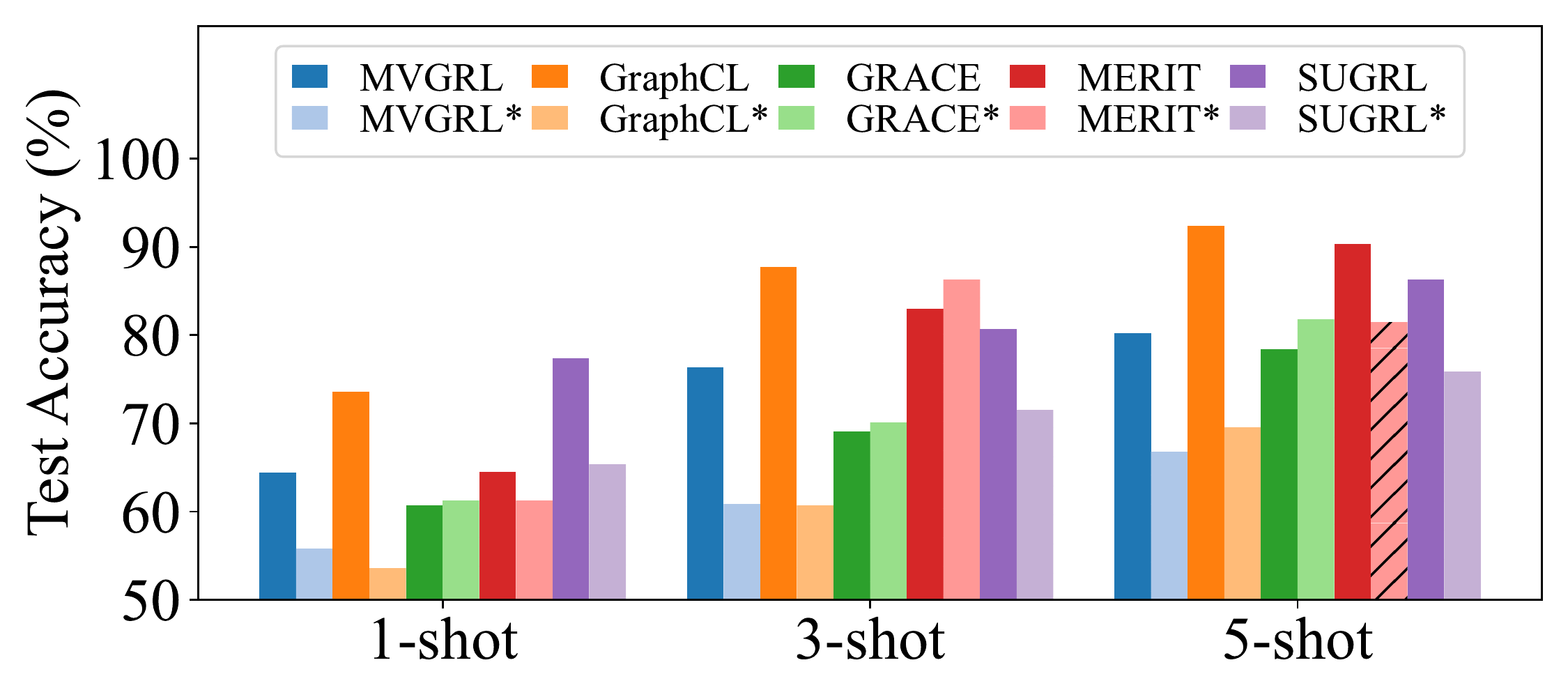}}

\caption{$2$-way $K$-shot results on \texttt{Amazon-Computer} and \texttt{CiteSeer}, TLP with self-supervised and supervised GCL. TLP Methods with $\ast$ are based on supervised GCL methods.}

	\end{figure}

% \newpage

% \newpage
% \section{More Discussions}
% \textcolor{black}{
% \subsection{Variations within each Category of Methods across Different Datasets}
% }
% \textcolor{black}{
% \subsection{Is TLP always better than Meta-learning?}
% }
% \textcolor{black}{
% \subsection{Convergence Time of Different Methods}
% }

\end{document}